\documentclass[runningheads]{llncs}

\usepackage{eccv}

\usepackage{eccvabbrv}

\usepackage{graphicx}
\usepackage{booktabs}
\usepackage{multirow}
\usepackage{pifont}

\usepackage{wrapfig}

\usepackage[dvipsnames]{xcolor}
\newcommand{\rf}[1]{{\color{red}#1}}
\newcommand{\bd}[1]{{\color{blue}#1}}

\usepackage[accsupp]{axessibility}  

\usepackage[breaklinks,colorlinks,citecolor=eccvblue]{hyperref}

\usepackage{orcidlink}

\begin{document}

\title{Collaborative Feedback Discriminative Propagation for Video Super-Resolution}

\titlerunning{CFD Propagation for Video Super-Resolution}

\author{Hao Li \and
Xiang Chen \and
Jiangxin Dong \and
Jinhui Tang \and
Jinshan Pan}

\authorrunning{H.~Li et al.}

\institute{Nanjing University of Science and Technology}

\maketitle

\vspace{-1.5em}
\begin{abstract}
The key success of existing video super-resolution (VSR) methods stems mainly from exploring spatial and temporal information, which is usually achieved by a recurrent propagation module with an alignment module.
However, inaccurate alignment usually leads to aligned features with significant artifacts, which will be accumulated during propagation and thus affect video restoration.
Moreover, propagation modules only propagate the same timestep features forward or backward that may fail in case of complex motion or occlusion, limiting their performance for high-quality frame restoration.
To address these issues, we propose a collaborative feedback discriminative (CFD) method to correct inaccurate aligned features and model long-range spatial and temporal information for better video reconstruction.
In detail, we develop a discriminative alignment correction (DAC) method to adaptively explore information and reduce the influences of the artifacts caused by inaccurate alignment.
Then, we propose a collaborative feedback propagation (CFP) module that employs feedback and gating mechanisms to better explore spatial and temporal information of different timestep features from forward and backward propagation simultaneously.
Finally, we embed the proposed DAC and CFP into commonly used VSR networks to verify the effectiveness of our method.
Quantitative and qualitative experiments on several benchmarks demonstrate that our method can improve the performance of existing VSR models while maintaining a lower model complexity.
The source code and pre-trained models will be available at \href{https://github.com/House-Leo/CFDVSR}{CFDVSR}.
\end{abstract}

\section{Introduction} \label{sec:introduction}
Video super-resolution (VSR) aims to reconstruct high-resolution (HR) video sequence from given low-resolution (LR) video sequence.
Compared to single image super-resolution (SISR) methods~\cite{dong2014learning,li2019feedback,haris2018deep,zhang2018image}, VSR methods not only rely on self-similarity within local spatial features to reconstruct high-frequency details but also need to model the temporal information from inter-frames to restore HR video sequence.
Thus, restoring HR video is quite challenging as the inaccurate spatio-temporal information extracted from degraded LR video.

Most existing deep convolutional neural networks (CNNs) based methods
\begin{wrapfigure}[19]{r}{0.6\textwidth}
	\centering
	\includegraphics[width=0.58\textwidth]{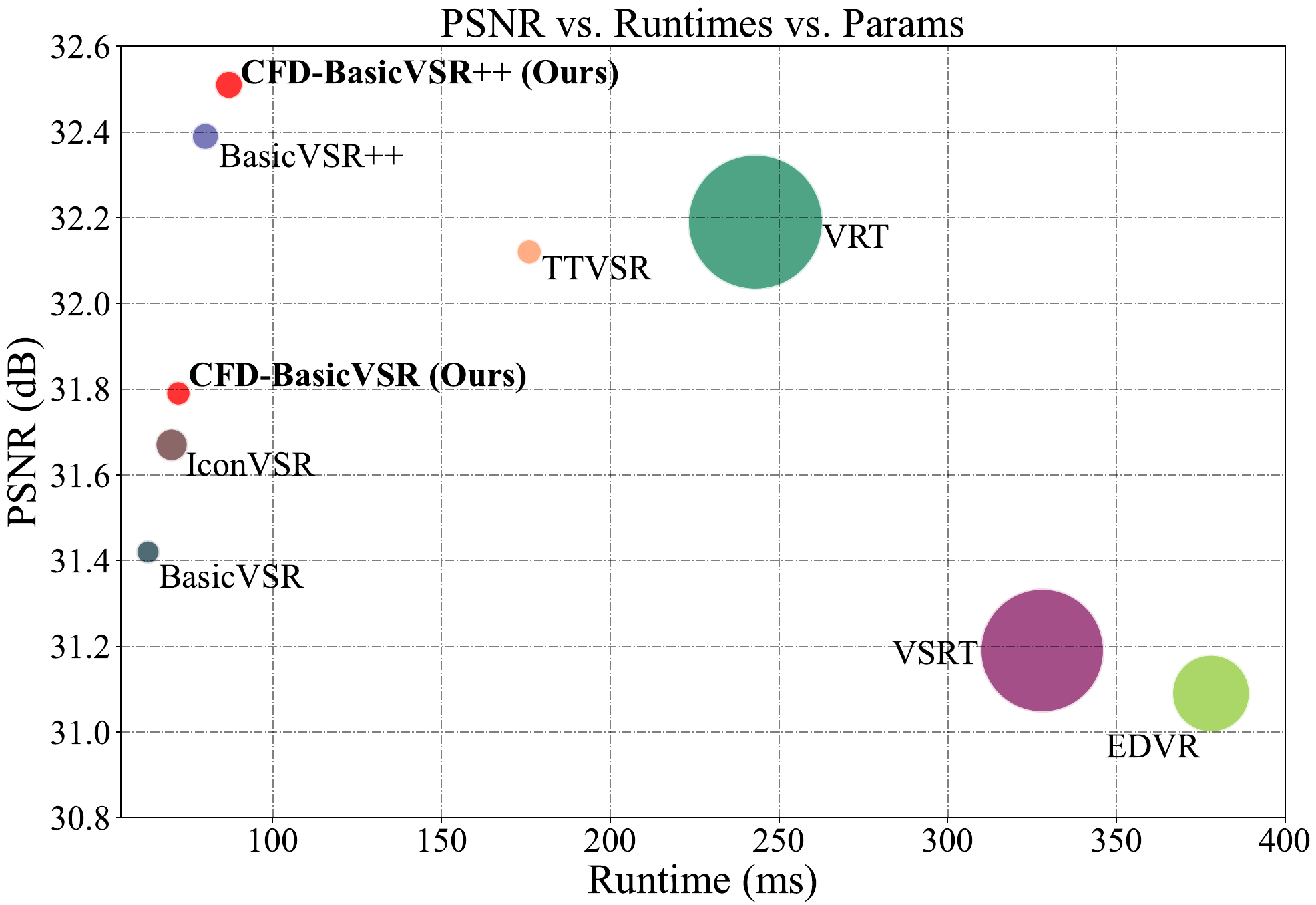}
	\vspace{-1mm}
	\caption{Comparison results of our proposed CFD-BasicVSR, CFD-BasicVSR++ and other methods on the REDS4 dataset~\cite{nah2019ntire} in terms of PSNR, running time and model parameters. Circle sizes indicate the number of parameters. Both our proposed models achieve a better trade-off between efficiency and performance.}
	\label{fig:comp}
	\vspace{-2mm}
\end{wrapfigure}
adopt a recurrent propagation block with an alignment module (\eg,  the deformable convolution~\cite{wang2019edvr,ying2020deformable,tian2020tdan} and optical flow~\cite{sajjadi2018frame,jo2018deep,tao2017detail,chan2021basicvsr,chan2022basicvsrplusplus,liang2022vrt,liu2022ttvsr}) to model spatio-temporal information from inter-frame, where the alignment module
%
%
%
%
%
is used to estimate the displacement between two adjacent frames for the motion compensation.
However, most of the displacement estimation methods are designed for high-quality images, directly using these approaches to low-resolution frames usually leads to inaccurate results, which accordingly leads to warped images or features with details information loss and severe artifacts.
These artifacts would accumulate when using the recurrent propagation to explore temporal information from long-range frames, thus affecting high-quality frame reconstruction.
Therefore, it is essential to reduce the influences of inaccurate warped features.

\vspace{-0.5mm}
We note that most recurrent propagation methods~\cite{chan2021basicvsr,sajjadi2018frame,isobe2020video} usually propagate video frames forward and backward independently, lacking the interaction of temporal information.
To address this issue, BasicVSR++~\cite{chan2022basicvsrplusplus} introduces the second-order grid connection into the bidirectional recurrent propagation module, which can refine features by aggregating features of the same timestep in different propagation branches.
However, if the features from current timestep are not accurately aligned, the above errors would be amplified due to the grid connection.
Therefore, it is of great interest to develop a propagation module with a correction method to discriminatively calibrate inaccurate aligned features, reduce errors during propagation, and bring more temporal interaction between different timestep features.

\vspace{-0.5mm}
To this end, we develop a collaborative feedback discriminative (CFD) propagation method to better explore the spatio-temporal information for VSR.
As the estimated optical flow from LR frames are not always accurate and may lead to the aligned results with significant artifacts, we develop a simple yet effective discriminative alignment correction (DAC) module to adaptively explore information to calibrate the inaccurate aligned features that suppress artifacts generation.
However, the recurrent propagation module with DAC does not effectively explore spatial and temporal information, as using only past or future information to compensate for the current frame may fail in complex motion case.
Thus, we propose a collaborative feedback propagation (CFP) module that utilizes the feedback and gating mechanisms to propagate different timestep features from backward and forward simultaneously for long-range temporal information exploration and better video reconstruction.

To demonstrate the effectiveness of our method, we incorporate our method into two prevalent CNN-based VSR backbones, BasicVSR~\cite{chan2021basicvsr} and BasicVSR++~\cite{chan2022basicvsrplusplus}, and a Transformer-based VSR backbnone PSRT~\cite{shi2022rethinking}, resulting in three redesigned networks denoted as CFD-BasicVSR, CFD-BasicVSR++, and CFD-PSRT, respectively.
Figure~\ref{fig:comp} illustrates that our redesigned models achieve favorable performance with a better trade-off between efficiency and performance.

The main contributions are summarized as follows:
\begin{itemize}
	\item We propose a simple yet effective discriminative alignment correction module to adaptively reduce the influcence of the artifacts caused by inaccurate alignment during feature warping.
	\item We develop a collaborative feedback propagation module based on the feedback and gating mechanisms to jointly propagate different timestep features from backward and forward branches for long-range spatio-temporal information exploration.
    \item We formulate both the discriminative alignment correction module and collaborative feedback propagation module into a unified model based on existing VSR network and show that our method can significantly improve the performance of existing VSR models but to not increase model complexity and computational cost significantly.
\end{itemize}

\section{Related Work}
\noindent \textbf{Video super-resolution.}
Unlike SISR, VSR relies heavily on the inter-frame sub-pixel and temporal information.
Previous VSR methods can be roughly divided into two categories: \textit{sliding-window} and \textit{recurrent propagation}.

\begin{figure*}[t]
	\centering
	\includegraphics[width=1.0\textwidth]{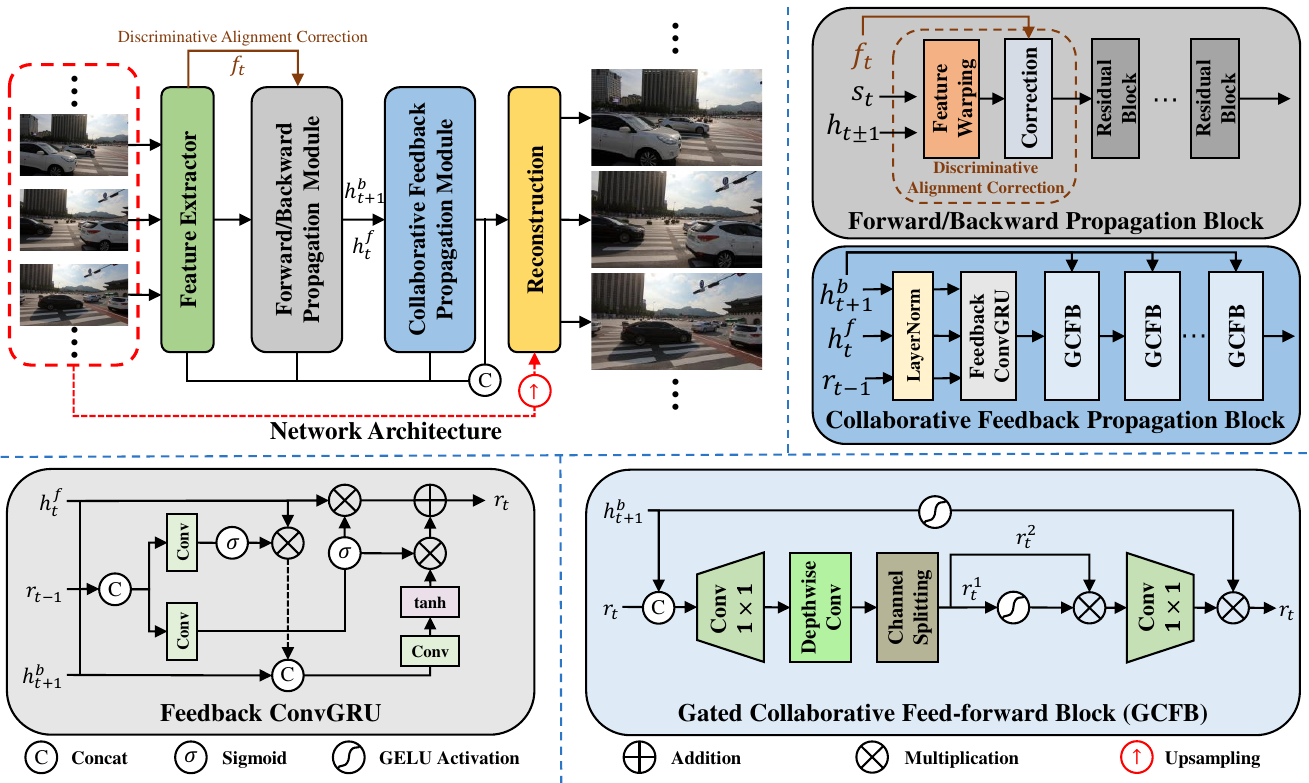}
	\caption{The overall architecture of our proposed model. Our model consists of a feature extractor, a forward/backward propagation module with the discriminative alignment correction (DAC), a collaborative feedback propagation (CFP) module, and reconstruction module. The DAC uses shallow features $f_{t}$ to explore more details information after feature warping, which corrects the aligned features for propagation. Moreover, as the core components of our CFP, Feedback ConvGRU and gated collaborate feed-forward block (GCFB) bring more temporal interactions between different timestep features from forward and backward propagation simultaneously. Here $s_{t}$ is the optical flow at $t$-th timestep, $h^{\{f,b\}}_{t}$ and $r_{t}$ denote the $t$-th timestep features in forward/backward propagation and CFP, respectively.}
	\vspace{-6mm}
	\label{fig:network}
\end{figure*}

\textit{Sliding-window} based methods usually employ optical flow or a learnable motion compensation layer to align neighboring frames within a short sequence video input.
SPMC~\cite{xue2019video} employs a sub-pixel motion compensation layer to simultaneously perform up-sampling and differentiable image warping.
EDVR~\cite{wang2018video} and TDAN~\cite{tian2020tdan} introduce deformable convolution for feature alignment.
PFNL~\cite{yi2019progressive} and MuCAN~\cite{li2020mucan} try to explore the infer-frame temporal correspondences based on a non-local aggregation module.
Although these methods achieve promising results, such an approach based on local information propagation inevitably increases runtime and computational costs, and methods with explicit motion compensation may produce artifacts due to inaccurate motion estimation.

\textit{Recurrent propagation} based methods exploit long-range temporal information using recurrent convolutional networks.
FRVSR~\cite{sajjadi2018frame} recurrently propagates the restored HR video frames to obtain the final HR outputs.
RSDN~\cite{isobe2020video} decomposes the input frame into structure and detail components and proposes a recurrent structure-detail block for unidirectional propagation.
In addition, it employs a hidden-state adaptation module to enhance the model robustness.
However, these methods based on unidirectional propagation only utilize the past estimated results to current features for better reconstruction.
To this end, BasicVSR~\cite{chan2021basicvsr} adopts a bidirectional temporal propagation that propagates video sequence forward and backward in time independently, and proposes flow-based feature alignment to achieve significant improvement.
BasicVSR++~\cite{chan2022basicvsrplusplus} modifies BasicVSR by employing second-order grid propagation and flow-guided deformable alignment, achieving state-of-the-art performance in VSR task.
In particular, the grid connection in BasicVSR++ enables repeated refinement which brings information interaction between forward and backward propagation at the same timestep.
Nevertheless, these methods rely on the accurate estimated motion, and if the current features are inaccurately aligned, the errors would be accumulated during the propagation, leading to noticeable artifacts in VSR.
Different from these works, we propose to calibrate the inaccurate aligned features before propagation for effectively suppressing the influence of artifacts and better video restoration.

\noindent \textbf{Feedback mechanism.} The feedback mechanism aims to aggregate the high-level features in the deep layer with low-level features in the shallow layer, which has been widely used in other low-level image tasks~\cite{li2019feedback,deng2021feedback,chen2021robust,li2019gated}.
DBPN~\cite{haris2018deep} uses a back-projection mechanism to achieve iterative error feedback through up- and down-projection units.
Li \textit{et al.}~\cite{li2019feedback} design a feedback block for processing the feedback information flow, effectively enhancing low-level representations with high-level ones.
Deng \textit{et al.}~\cite{deng2021feedback} propose a coupled feedback network that consists of two coupled recursive sub-networks, to refine the fused HR image progressively.
However, since the bidirectional recurrent propagation employs forward and backward propagation following the temporal order, there is few work applying feedback mechanism in VSR task.
The most relevant work to feedback learning using in VSR task is ConvLSTM~\cite{zamir2017feedback}.
Xiang \textit{et al.}~\cite{xiang2020zooming} propose an end-to-end framework that uses deformable ConsLSTM to process the temporally consecutive features for space-time video super-resolution (STVSR).
Lin \textit{et al.}~\cite{seq2seq} design a sequence-to-sequence model for video restoration, which employs stacked ConvGRU blocks modified by ConLSTM to model the long-term temporal dependencies.
However, these methods do not consider using the hidden state in future features to refine the current features, which is similar to the idea of feedback mechanisms.
Inspired by that, we propose a collaborative feedback propagation module to model long-range spatio-temporal information for VSR.
With our proposed module, information from future features can feedback to the current features for comprehensive refinement.
To the best of our knowledge, our work is the first attempt to employ feedback mechanism in VSR task.

\section{Proposed Method}
To reduce the influence of the artifacts caused by inaccurate feature alignment and better explore spatio-temporal information, we develop a collaborative feedback discriminative (CFD) propagation method that includes a discriminative alignment correction (DAC) and a collaborative feedback propagation (CFP) module (see Figure~\ref{fig:network} for the detailed network architecture).
The DAC is used to adaptively calibrate the inaccurate aligned features, while the CFP learns spatio-temporal information by jointly propagating the backward and forward features for better video reconstruction.
In the following, we present the details of our method and how to integrate it into existing models.

\subsection{Discriminative alignment correction}
Existing methods~\cite{kim2018spatio,chan2021basicvsr,liu2022ttvsr,liang2022vrt} have demonstrated that optical flow based alignment effectively leverages inter-frame temporal information, leading to improved performance of VSR models.
For example, given the adjacent LR video frames $\{I_{t},I_{t \pm 1}\} \in \mathbb{R}^{3 \times H \times W}$, corresponding to the timestep $\{t,t \pm 1\}$, feature alignment method first computes forward and backward optical flow using the pre-trained models (\eg, SpyNet~\cite{ranjan2017optical}, PWCNet~\cite{Sun2018PWC-Net}). Then it uses the spatial warping module $\mathcal{W}(\cdot)$ to process the propagation features $\{h^{f}_{t - 1},h^{b}_{t + 1}\} \in \mathbb{R}^{C \times H \times W}$, resulting in the warped features $\{\hat{h}_{t}^{f}, \hat{h}_{t}^{b}\}$, which are formulated as:
\begin{equation}
    \begin{split}
        \{s_{t}^{f},s_{t}^{b}\} &= \mathcal{S}(I_{t},I_{t \pm 1}), \\
        \hat{h}_{t}^{f} &= \mathcal{W}(h^{f}_{t - 1}\,s_{t}^{f}), \\
        \hat{h}_{t}^{b} &= \mathcal{W}(h^{b}_{t + 1}\,s_{t}^{b}),
    \end{split}
\end{equation}
where $\mathcal{S}(\cdot)$ denotes the optical flow estimation module; $s_{t}^{f}$ and $s_{t}^{b}$ are the forward and backward optical flow.

\begin{figure}[t]
\scriptsize
\centering
    \begin{tabular}{ccc}
      \includegraphics[width=0.28\textwidth]{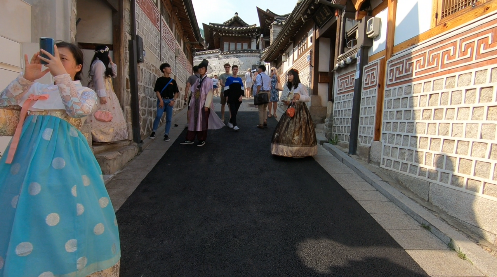}
    & \includegraphics[width=0.28\textwidth]{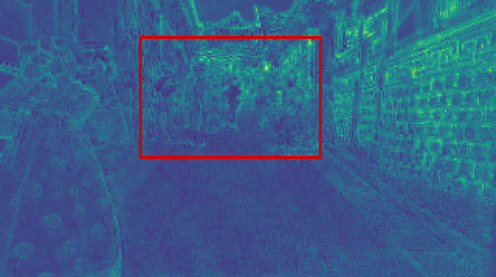}
    &  \includegraphics[width=0.28\textwidth]{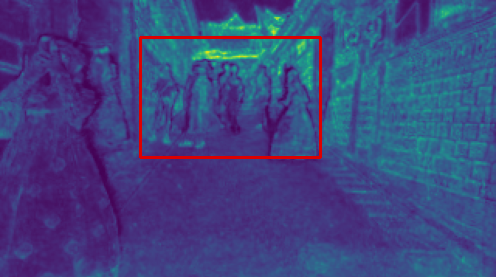}   \\
    (a) Reference frame                           &  (c) Aligned by feature alignment  & (e) Aligned by DAC \\
       \includegraphics[width=0.28\textwidth]{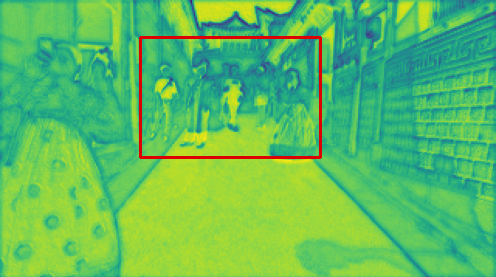}
     & \includegraphics[width=0.28\textwidth]{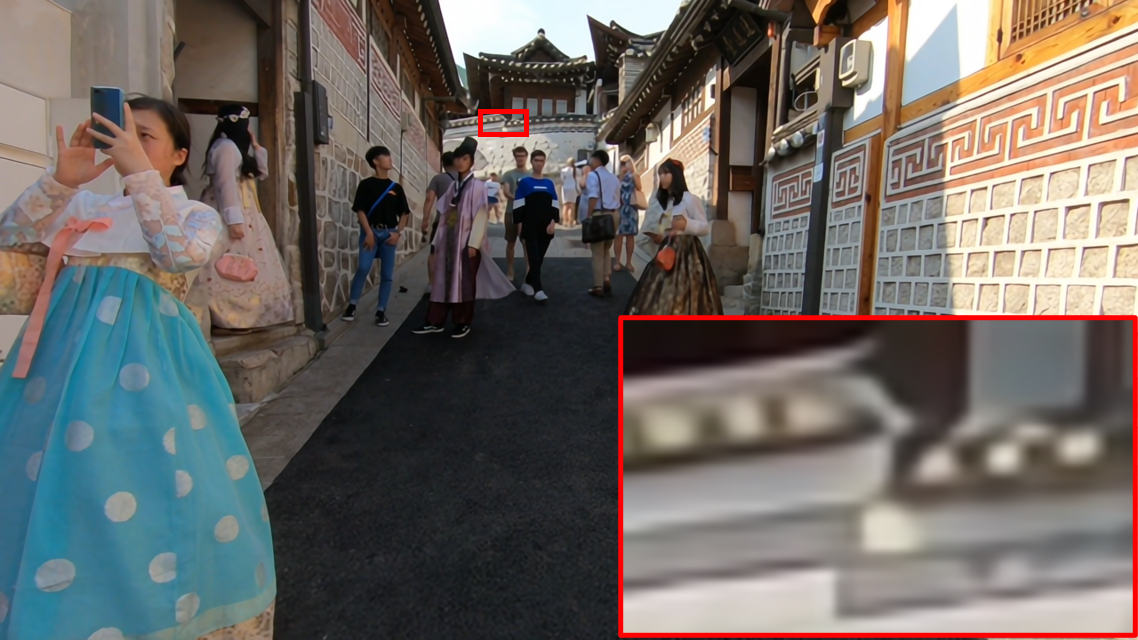}
    & \includegraphics[width=0.28\textwidth]{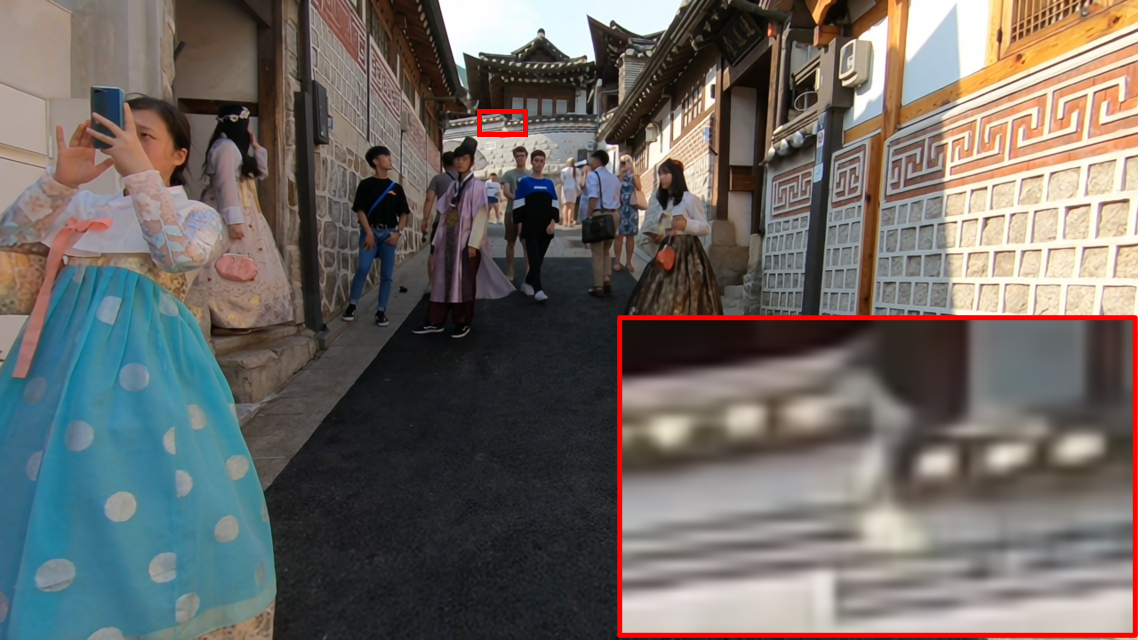}     \\
    (b) Shallow features & (d) VSR result with (c)                       &  (f) VSR result with (e)
    \end{tabular}
    \caption{\textbf{Effect of the proposed DAC method on VSR.} The inaccurate flow estimation and the resampling operation in the spatial warping module cause severe damage to structure and edge information, which is prone to generate artifacts in corresponding regions during feature alignment (see \rf{red} box in (c)). In contrast, we notice that the shallow features (b) could better preserve such information. This instinctively prompts us to compensate for the information loss using shallow features. With our proposed DAC method, the aligned features can accurately recover the information in this region (see \rf{red} box in (d)), and the restored VSR result (f) has sharper texture details than (d).
    }
    \label{fig:alignment}
\vspace{-6mm}
\end{figure}

However, if the optical flow is not estimated accurately, it would lead to the warped features with significant artifacts, which interferes with the main structures and details. For example, the contours of pedestrians and buildings are not recognized as shown in Figure~\ref{fig:alignment}(c).
As point out by~\cite{Pan_2023_CVPR}, the errors caused by the inaccurate alignment will accumulate and thus affect the clear frame restoration.
For example, the structures of buildings are not restored well as shown in Figure~\ref{fig:alignment}(d).
Thus, reducing the influences of artifacts in the warped features is important for VSR.


%
%

As most useful details and structures (\eg, object boundaries) usually higher response values in the extracted features, while the artifacts caused by the inaccurate alignment usually reduce the response values.
Simply using the warped features with artifacts in the propagation will increase the difficulty of the network for the high-frequency information estimation.
%
%

We note that an ideal warped feature should align with the reference feature well.
As the shallow features extracted from reference images contain high-frequency information, \eg, richer details and structures (see Figure~\ref{fig:alignment}(b)), we use them to correct the errors caused by the artifacts in the warped features.

As the regions of a feature extracted by CNNs with high response values usually corresponds to absolute intensity values, we develop a DAC module to enable the network to estimate features with finer details and structures:
%
%
%

%
\begin{equation}
     h_{t}^{\{f,b\}}(\mathrm{x}) =  \left\{
                                        \begin{aligned}
                                            & \hat{h}_{t}^{\{f,b\}}(\mathrm{x}),~ |\hat{h}_{t}^{\{f,b\}}(\mathrm{x})| \geq |f_{t}(\mathrm{x})|, \\
                                            & f_{t}(\mathrm{x}),~~~~~~~|\hat{h}_{t}^{\{f,b\}}(\mathrm{x})|<|f_{t}(\mathrm{x})|,
                                        \end{aligned}
                                    \right.
    \label{eq:dac}
\end{equation}
where $\mathrm{x}$ is the coordinate of pixel, $h_{t}^{\{f,b\}}$ denotes the enhanced warped features for forward and backward propagation.
Since features at the same timestep have the most similar contents, we use the $t$-th timestep features extracted by the feature extractor (see Figure~\ref{fig:network}) as the shallow features $f_{t}$, to adaptively correct the artifacts in the aligned features by iterative learning of the feature extractor.
%
%
Eq.~\eqref{eq:dac} can be regarded as a special feature correction operation that uses a feature with rich high-frequency information as a guidance to recover the high response regions destroyed during feature warping.
%
The following propagation module will further extract the hidden features $h_{t}^{\{f,b\}}$.

Unlike the previous alignment methods, our DAC can restore more realistic image details (see Figure~\ref{fig:alignment}(f)) and significantly improve the performance of the model.
We will show its effectiveness in Section~\ref{sec:analy}.

\subsection{Collaborative feedback propagation module}
The recurrent propagation module with our DAC only uses past or future information to compensate for the current features that may fail in some complex motion or occlusion cases.
An intuitive idea is to use more different timestep information for comprehensive refinement in response to various complex cases.
Recent work~\cite{isobe2022look} attempts to exploit intermediate HR results in both the future and past to refine the current output.
However, this operation in HR space requires higher computational costs, and~\cite{isobe2022look} is a uni-directional architecture limiting its performance in VSR.

To solve this issue, we develop a collaborative feedback propagation module that explores future and past information to comprehensively refine the current features in LR space.
Our CFP is inspired by the success of feedback learning in image SR~\cite{li2019feedback,deng2021feedback} and ConvGRU in VSR~\cite{xiang2020zooming,seq2seq}.
It consists of a LayerNorm~\cite{ba2016layer} layer, a feedback ConvGRU block and a stacking of Gated Collaborative Feed-forward Blocks (GCFBs) that aggregate different timestep features from forward and backward propagation.

\noindent \textbf{Feedback ConvGRU.} The Feedback ConvGRU is mainly used to extract and interact the temporal information from different timestep input features.
Compared with previous ConvGRU~\cite{zamir2017feedback,xiang2020zooming} methods only consider past information, the proposed Feedback ConvGRU considers more temporal information and fully utilizes hidden features from each propagation branch.
Specifically, we get the hidden features $\{h_{t}^{f},h_{t+1}^{b}\}$ from the forward and backward propagation.
Then, we take $\{h_{t}^{f}$, $h_{t+1}^{b}\}$ and the past frame features $r_{t-1}$ as the feedback ConvGRU block input, the entire procedure is formulated as:
\begin{equation}
    \begin{split}
        & v = \text{LN}(\text{Concat}(r_{t-1},h_{t}^{f},h_{t+1}^{b})), \\
        & z = \sigma(\text{Conv}(v)),~~w = \sigma(\text{Conv}(v)), \\
        & q = \text{tanh}(\text{Conv}(\text{Concat}(w \otimes h_{t}^{f}, h_{t+1}^{b}))), \\
        & r_{t} = (1-z) \otimes h_{t}^{f} + z \otimes q,
    \end{split}
    \label{eq:GRU}
\end{equation}
where $\text{Conv}(\cdot)$ is the $3 \times 3$ convolution, $\text{LN}(\cdot)$ and $\text{Concat}(\cdot)$ denote the layer normalization and concatenation operation, $\sigma(\cdot)$ and $\otimes$ are the sigmoid function and the element-wise multiplication, respectively, and $r_{t-1}=r_0$ is the initiation propagation features when $t=1$.

\noindent \textbf{Gated collaborative feed-forward block.} Note that the Feedback ConvGRU only considers the temporal information between three timestep features (\ie, $h_{t}^{f}, h_{t+1}^{b}$ and $r_{t-1}$).
To better explore spatial information and keep temporal information interaction, we develop the GCFB to refine the output $r_t$ from Feedback ConvGRU.
Since Eq.~\eqref{eq:GRU} uses the gating mechanism on the forward propagation features $h_{t}^{f}$, the results $r_t$ has more information coming from forward propagation branch.
Thus, we use the backward propagation features $h_{t+1}^{b}$ for gating in the following GCFBs so that $r_t$ has information from different propagation branch.
%
%
Specifically, we first concatenate $r_t$ and $h_{t+1}^{b}$ as input and use a $1 \times 1$ convolution to expand the channel.
Then, we apply a depth-wise convolution to extract the spatial information and two gated units to achieve temporal interaction.
%
Given the input features $r_t \in \mathbb{R}^{C \times H \times W}$ and $h_{t+1}^{b}$, this procedure can be expressed as:
\begin{equation}
    \begin{split}
        & \hat{r}_t = \text{Conv}_{1 \times 1}(\text{Concat}(r_t,h_{t+1}^{b})),~\hat{r}_t \in \mathbb{R}^{4C \times H \times W}, \\
        & \{r^1_{t},r^2_{t}\} = \text{Split}(\text{DConv}(\hat{r}_t)),~\{r^1_{t},r^2_{t}\} \in \mathbb{R}^{2C \times H \times W}, \\
        & \overline{r}_t = \phi(r^1_{t}) \otimes r^2_{t},~\overline{r}_t \in \mathbb{R}^{2C \times H \times W},\\
        & r_t = \text{Conv}_{1 \times 1}(\overline{r}_t) \otimes \phi(h_{t+1}^{b}),~r_t \in \mathbb{R}^{C \times H \times W},
    \end{split}
    \label{eq:GCFB}
\end{equation}
where $\phi$ represents the GELU function and $\otimes$ denotes the element-wise multiplication, $\text{Split}(\cdot)$ is the channel split operation, $\text{Conv}_{1 \times 1}(\cdot)$ and $\text{DConv}(\cdot)$ denote the $1 \times 1$ convolution and the depth-wise convolution with a kernel size of $3 \times 3$ pixels, respectively.
%
%
The first gating mechansim in Eq.~\eqref{eq:GCFB} is the self-gating, while the second one is the gating of future features, enabling $r_t$ to learn information from both past and future features.
Different from the existing gating method~\cite{Zamir2021Restormer} that causes pixel misalignment by using the residual connections, our proposed GCFB can more effectively introduce temporal interactions among features at different timesteps, which is crucial for VSR.
%

\subsection{CFD for VSR backbones}
As shown in Figure~\ref{fig:network}, we design a new VSR architecture based on our proposed CFD propagation method, combined with the existing VSR backbones including recurrent propagation.
We first describe the overall pipeline of our network architecture for VSR, and then introduce three new VSR backbones, CFD-BasicVSR, CFD-BasicVSR++, and CFD-PSRT.

\noindent \textbf{Overall pipeline.} Specifically, given LR video sequence $\mathbf{I}^{L}=\{I^{L}_t\}_{t=1}^{N}$ with $N$ frames, we first use a feature extractor which contains a convolution layer and five Resblocks~\cite{he2016deep} to extract shallow features $\mathbf{F}=\{f_{t}\}^N_{t=1}$ from each input frame.
Taking $\mathbf{F}$ as input, we utilize the forward/backward propagation module with discriminative alignment correction to explore spatio-temporal information by correcting artifacts generated by inaccurate aligned features, to obtain the forward and backward propagation features $\mathbf{H}=\{h^{\{f,b\}}_{t}\}^N_{t=1}$.
In addition, to better exploit long-range spatio-temporal information from different timestep features, we utilize our CFP to process $\mathbf{H}$ and get the refined features $\mathbf{R}=\{r_{t}\}^N_{t=1}$.
Finally, we restore the HR video sequence $\mathbf{I}^{H}=\{I^{H}_{t}\}^N_{t=1}$ by aggregating the shallow, propagation and refine features using a reconstruction module $\mathcal{R}(\cdot)$ which contains five Resblocks~\cite{he2016deep} and two pixel-shuffle layers~\cite{shi2016real}. For $t$-th video frame, the reconstructed HR frame $I_{t}^H$ which is formulated as:
\begin{equation}
    \begin{split}
        I^{H}_{t} = \mathcal{R}(\text{Concat}(f_t, h_t^f, h_t^b, r_t)) + I^L_{t}\uparrow,
    \end{split}
\end{equation}
where $\mathcal{R}(\cdot)$ denotes the reconstruction module, $\uparrow$ represents the upsampling operation, and $\{f_t, h_t^f, h_t^b, r_t\}$ are the features from feature extractor, forward propagation, backward propagation and CFP, respectively.

\noindent \textbf{CFD-BasicVSR, CFD-BasicVSR++ and CFD-PSRT.}
To demonstrate the effectiveness of our proposed method, we choose three strong VSR backbones as the base models and add our CFD propagation method into them, including two CNN-based models (\ie, BasicVSR~\cite{chan2021basicvsr} and BasicVSR++~\cite{chan2022basicvsrplusplus}) and a Transformer-based model (\ie, PSRT~\cite{shi2022rethinking}).
We denote these three new models as CFD-BasicVSR, CFD-BasciVSR++, and CFD-PSRT, respectively.
The quantitative results of our models are presented in Table~\ref{tab:quan}.


\begin{table*}[!t]
    \caption{Quantitative comparisons with state-of-the-art methods for VSR ($\times 4$). \rf{Red} and \bd{Blue} indicate the best and the second-best performance, respectively. The runtime is the average inference time on 100 LR video frames with a size of $180{\times}320$ resolution. All results are calculated on Y-channel except REDS4~\cite{nah2019ntire} (RGB-channel).}
    \label{tab:quan}
    \vspace{-2mm}
    \centering
    \resizebox{1.0\textwidth}{!}{
            \begin{tabular}{l||c|c|c||c|c|c||c|c}
                \toprule
                \multirow{2}{*}{Methods}           & \multicolumn{3}{c||}{BI degradation}                                                                    & \multicolumn{3}{c||}{BD degradation}              & \multirow{2}{*}{Params (M)}            & \multirow{2}{*}{Runtime (ms)}   \\ \cline{2-7}
                                                   & REDS4~\cite{nah2019ntire}            & Vimeo-T~\cite{xue2019video}      & Vid4~\cite{liu2014bayesian} & UDM10~\cite{yi2019progressive} & Vimeo-T~\cite{xue2019video}   & Vid4~\cite{liu2014bayesian}     &            &         \\ \hline \hline
                Bicubic                            & 26.14/0.7292                         & 31.32/0.8684                       & 23.78/0.6347                & 28.47/0.8253                   & 31.30/0.8687                    & 21.80/0.5246                    & -          & -       \\
                TOFlow~\cite{xue2019video}         & 27.98/0.7990                         & 33.08/0.9054                       & 25.89/0.7651                & 36.26/0.9438                   & 34.62/0.9212                    & -                               & -          & -       \\
                FRVSR~\cite{sajjadi2018frame}      & -                                    & -                                  & -                           & 37.09/0.9522                   & 35.64/0.9319                    & 26.69/0.8103                    & 5.1        & 137     \\
                DUF~\cite{jo2018deep}              & 28.63/0.8251                         & -                                  & -                           & 38.48/0.9605                   & 36.87/0.9447                    & 27.38/0.8329                    & 5.8        & 974     \\
                RBPN~\cite{haris2019recurrent}     & 30.09/0.8590                         & 37.07/0.9435                       & 27.12/0.8180                & 38.66/0.9596                   & 37.20/0.9458                    & -                               & 12.2       & 1507    \\
                EDVR~\cite{wang2019edvr}           & 31.09/0.8800                         & 37.61/0.9489                       & 27.35/0.8264                & 39.89/0.9686                   & 37.81/0.9523                    & 27.85/0.8503                    & 20.6       & 378     \\
                PFNL~\cite{yi2019progressive}      & 29.63/0.8502                         & 36.14/0.9363                       & 26.73/0.8029                & 38.74/0.9627                   & -                               & 27.16/0.8355                    & 3.0        & 295     \\
                MuCAN~\cite{li2020mucan}           & 30.88/0.8750                         & 37.32/0.9465                       & -                           & -                              & -                               & -                               & -          & -       \\
                BasicVSR~\cite{chan2021basicvsr}   & 31.42/0.8909                         & 37.18/0.9450                       & 27.24/0.8251                & 39.96/0.9694                   & 37.53/0.9498                    & 27.96/0.8553                    & 6.3        & 63      \\
                IconVSR~\cite{chan2021basicvsr}    & 31.67/0.8948                         & 37.47/0.9476                       & 27.39/0.8279                & 40.03/0.9694                   & 37.84/0.9524                    & 28.04/0.8570                    & 8.7        & 70      \\
                ETDM~\cite{isobe2022look}          & 32.15/0.9024                         & -                                  & -                           & 40.11/0.9707                   & -                               & 28.81/0.8725                    & 8.4        & 70      \\
                BasicVSR++~\cite{chan2022basicvsrplusplus}    & \bd{32.39}/\bd{0.9069}    & \bd{37.79/0.9500}                       & \bd{27.79/0.8400}                & \bd{40.72/0.9722}                   & \bd{38.21/0.9550}                    & \bd{29.04/0.8753}                    & 7.3        & 77      \\ \hline
                \textbf{CFD-BasicVSR}              & 31.79/0.8953                         & 37.55/0.9478                       & 27.47/0.8280                & 40.21/0.9695                   & 37.95/0.9527                    & 28.30/0.8597                    & 6.6        & 72      \\
                \textbf{CFD-BasicVSR++}            & \rf{32.51}/\rf{0.9083}               & \rf{37.90}/\rf{0.9504}             & \rf{27.84}/\rf{0.8406}      & \rf{40.77}/\rf{0.9726}         & \rf{38.36}/\rf{0.9557}          & \rf{29.14}/\rf{0.8760}          & 7.5        & 87      \\ \hline \hline
                VSRT~\cite{cao2021vsrt}            & 31.19/0.8815                         & 37.71/0.9494                       & 27.36/0.8258                & -                              & -                               & -                               & 32.6       & 328     \\
                VRT~\cite{liang2022vrt}            & 32.19/0.9006                         & 38.20/0.9530             & 27.93/0.8425      & -         & -               & -               & 35.6       & 243     \\
                TTVSR~\cite{liu2022ttvsr}          & 32.12/0.9021                         & -                                  & -                           & -                   & -                    & -                   & 6.8        & 176     \\
                RVRT~\cite{liang2022rvrt} & \bd{32.75/0.9113} & 38.15/0.9527 & 27.99/0.8462 & - & - & - & 10.8 & 183 \\
                PSRT~\cite{shi2022rethinking} & 32.72/0.9106 & \bd{38.27/0.9536} & 	\bd{28.07/0.8485} & - & - & - & 13.4 & 812
                \\ \hline
                 \textbf{CFD-PSRT}            & \rf{32.83/0.9140}   & \rf{38.33/0.9548} & \rf{28.18/0.8503} &- &- &-       & 13.6        & 820\\
                \bottomrule
            \end{tabular}}
        \vspace{-4mm}
\end{table*}

\section{Experimental Results}
\subsection{Datasets and parameter settings}
\noindent \textbf{Training datasets and benchmarks.} Following~\cite{liang2022vrt,chan2021basicvsr,chan2022basicvsrplusplus}, we use the commonly used VSR datasets, including REDS~\cite{nah2019ntire} and Vimeo-90K~\cite{xue2019video} for training.
To evaluate our method, we use REDS4\footnote{Clips 000, 011, 015, 020 of REDS.}, Vid4~\cite{liu2014bayesian}, Vimeo-T~\cite{xue2019video} and UDM10~\cite{yi2019progressive} as benchmarks, including two degradation - Bicubic (BI) and Blur Downsampling (BD).
All models are trained and tested for $4 \times$ super-resolution, and we adopt PSNR and SSIM as the evaluation metrics.

\begin{figure*}[t]
	\footnotesize
	\begin{center}
		\begin{tabular}{c c c c c c c c}
			\multicolumn{3}{c}{\multirow{5}*[34pt]{
             \includegraphics[width=0.49\linewidth,height=0.28\linewidth]{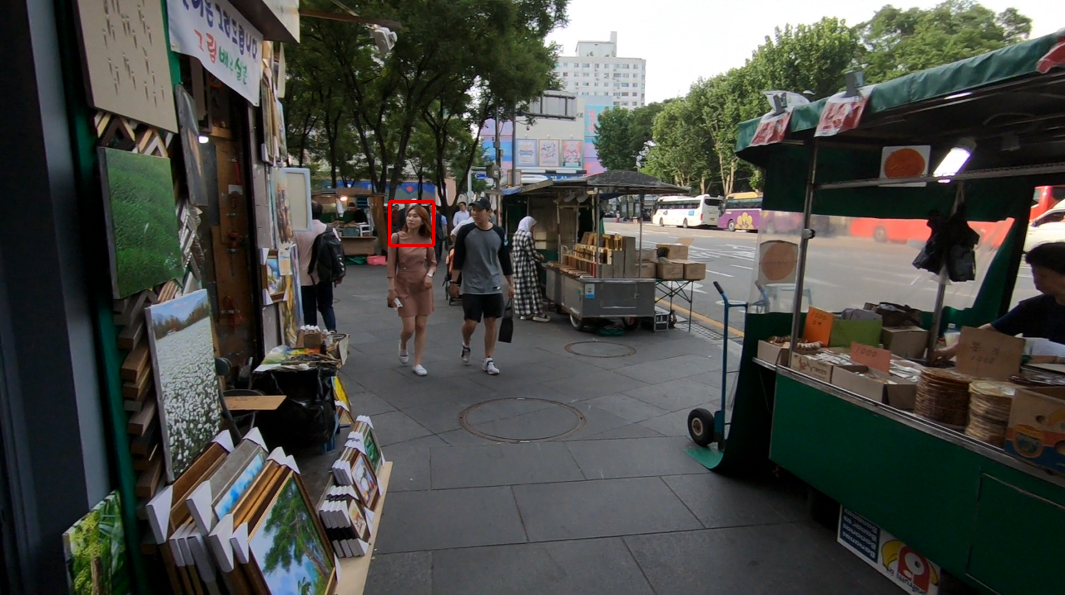}}}
            &  \includegraphics[width=0.12\linewidth]{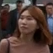}
            &  \includegraphics[width=0.12\linewidth]{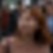}
            &   \includegraphics[width=0.12\linewidth]{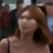}
            &  \includegraphics[width=0.12\linewidth]{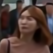}  \\
			\multicolumn{3}{c}{~}                                   & (a)                &  (b)                     & (c)   &  (d)  \\		
			\multicolumn{3}{c}{~}
            &  \includegraphics[width=0.12\linewidth]{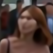}
            &  \includegraphics[width=0.12\linewidth]{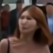}
            & \includegraphics[width=0.12\linewidth]{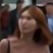}
            &  \includegraphics[width=0.12\linewidth]{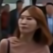}  \\
			\multicolumn{3}{c}{Frame 010, Clip 020} & (e)  &  (f)   &  (g)           & (h)  \\			
		\end{tabular}
	\end{center}
	\vspace{-6mm}
	\caption{Visual comparisons on the REDS4 dataset. (a) Ground truth. (b)-(h) denote the results generated by Bicubic, BasicVSR~\cite{chan2021basicvsr}, BasicVSR++~\cite{chan2022basicvsrplusplus}, VRT~\cite{liang2022vrt}, TTVSR~\cite{liu2022ttvsr}, CFD-BasicVSR (Ours), and CFD-BasicVSR++ (Ours), respectively. The results in (b)-(f) do not have accurate details information of eyes and mouths. However, the proposed methods (g) and (f) can effectively restore the facial details.}
	\label{fig:visual3}
	\vspace{-2mm}
\end{figure*}

\begin{figure*}[!t]
	\footnotesize
	\begin{center}
		\begin{tabular}{c c c c c c c c}
			\multicolumn{3}{c}{\multirow{5}*[34pt]{
             \includegraphics[width=0.49\linewidth,height=0.28\linewidth]{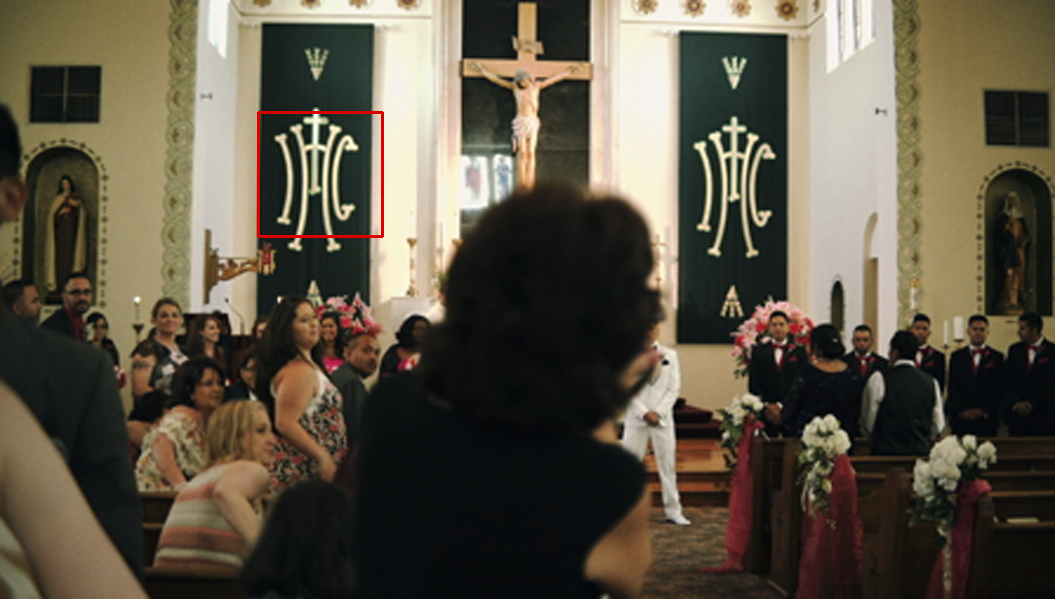}}}
            &  \includegraphics[width=0.12\linewidth]{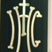}
            &  \includegraphics[width=0.12\linewidth]{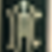}
            &  \includegraphics[width=0.12\linewidth]{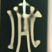}
            & \includegraphics[width=0.12\linewidth]{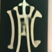}  \\
			\multicolumn{3}{c}{~}                                      &  (a)                 &  (b)                   &  (c)  & (d)  \\
			\multicolumn{3}{c}{~}
            &  \includegraphics[width=0.12\linewidth]{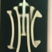}
            & \includegraphics[width=0.12\linewidth]{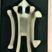}
            & \includegraphics[width=0.12\linewidth]{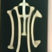}
            &  \includegraphics[width=0.12\linewidth]{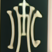}  \\
			\multicolumn{3}{c}{Sequence 837, Clip 001} &  (e)  &  (f)   & (g)      & (h)  \\			
		\end{tabular}
	\end{center}
	\vspace{-6mm}
	\caption{Visual comparisons on the Vimeo-T dataset. (a) Ground truth. (b)-(h) denote the results generated by Bicubic, BasicVSR~\cite{chan2021basicvsr}, BasicVSR++~\cite{chan2022basicvsrplusplus}, VRT~\cite{liang2022vrt}, TTVSR~\cite{liu2022ttvsr}, CFD-BasicVSR (Ours), and CFD-BasicVSR++ (Ours), respectively. The results restored by~\cite{chan2021basicvsr,chan2022basicvsrplusplus,liang2022vrt,liu2022ttvsr} still contain blurring and artifacts. In contrast, the proposed methods are able to accurately restore the upper-right part of the frame.}
	\label{fig:visual4}
	\vspace{-6mm}
\end{figure*}

\noindent \textbf{Parameters settings.} During the training, we use Adam optimizer~\cite{kingma2014adam} with $[\beta_1,\beta_2]=[0.9,0.99]$ and cosine annealing scheme~\cite{loshchilov2016sgdr}.
The batch size is set to be 16 and the patch size of input LR frames is $256 \times 256$, and the feature number $C$ is 64.
We follow the protocols of~\cite{chan2021basicvsr,chan2022basicvsrplusplus}, the initial learning rates of CFD-BasicVSR, CFD-BasicVSR++, and CFD-PSRT model are set to $2 \times 10^{-4}$, $1 \times 10^{-4}$, and $1 \times 10^{-4}$, respectively.
We use the pre-trained SpyNet~\cite{ranjan2017optical} and the weights of the flow estimator are fixed during the first 5,000 iterations with the initial learning rate $2.5 \times 10^{-5}$.
We also use a Resblock \cite{he2016deep} to mitigate the artifacts due to the destroyed structures or details after feature warping.
The total number of iterations of CFD-BasicVSR, CFD-BasicVSR++, and CFD-PSRT are set to be 300,000 600,000, and 600,000, respectively.
The number of GCFBs in CFP module is set to 5.
To constrain the network training, we use Charbonnier~\cite{charbonnier1994two} and FFT loss~\cite{cho2021rethinking} for training.
Furthermore, the number of recurrent propagation is set to be 2, which means that the network has twice forward and backward propagation.
All experiments are conducted with the PyTorch framework on 8 NVIDIA GeForce RTX 3090 GPUs.
Due to the page limit, more experimental settings and the detail of the network architecture are provided in the supplementary material.
The training code and pretrained models will be available to \href{https://github.com/House-Leo/CFDVSR}{CFDVSR}.

\vspace{-3mm}
\subsection{Comparisons with the state of the art}
\vspace{-1mm}
\label{sec:comparison}
To evaluate the performance of our proposed models, we conduct quantitative and qualitative experiments by comparing them with state-of-the-art VSR methods, including TOFlow~\cite{xue2019video}, FRVSR~\cite{sajjadi2018frame}, DUF~\cite{jo2018deep}, RBPN~\cite{haris2019recurrent}, EDVR~\cite{wang2019edvr}, PFNL~\cite{yi2019progressive}, MuCAN~\cite{li2020mucan}, BasicVSR~\cite{chan2021basicvsr}, IconVSR~\cite{chan2021basicvsr}, BasicVSR++~\cite{chan2022basicvsrplusplus}, ETDM~\cite{isobe2022look}, VSRT~\cite{cao2021vsrt}, VRT~\cite{liang2022vrt}, TTVSR~\cite{liu2022ttvsr}, RVRT~\cite{liang2022rvrt}, and PSRT~\cite{shi2022rethinking}.

\noindent \textbf{Quantitative comparisons.} Table~\ref{tab:quan} presents the quantitative results, compared with BasicVSR, our CFD-BasicVSR achieves significant improvements of 0.37dB and 0.42dB on the REDS4 and Vimeo-T(BD), respectively.
Furthermore, our CFD-BasicVSR++ achieves state-of-the-art performance on the REDS4 and the second-best results on other benchmarks.
Even compared to Transformer-based methods (\ie, VSRT, VRT and TTVSR), our approaches show comparable results.
Notably, CFD-BasicVSR++ outperforms VSRT on three BI degradation benchmarks, with only 23\% parameters.
Due to the self-attention and larger model complexity, VRT is better at handling short video sequence with higher performance on Vimeo-T (7 frames), Vid4 ($<$ 50 frames) and UDM10 (32 frames).
However, on the REDS4 dataset, CFD-BasicVSR++ exhibits a 0.32dB improvement in PSNR compared to VRT, with only 21\% parameters, indicating that our method can better handle long video sequence.

\begin{figure*}[t]
	\footnotesize
	\begin{center}
		\begin{tabular}{c c c c c c c}
			\multicolumn{3}{c}{\multirow{5}*[44pt]{
             \includegraphics[width=0.48\linewidth,height=0.34\textwidth]{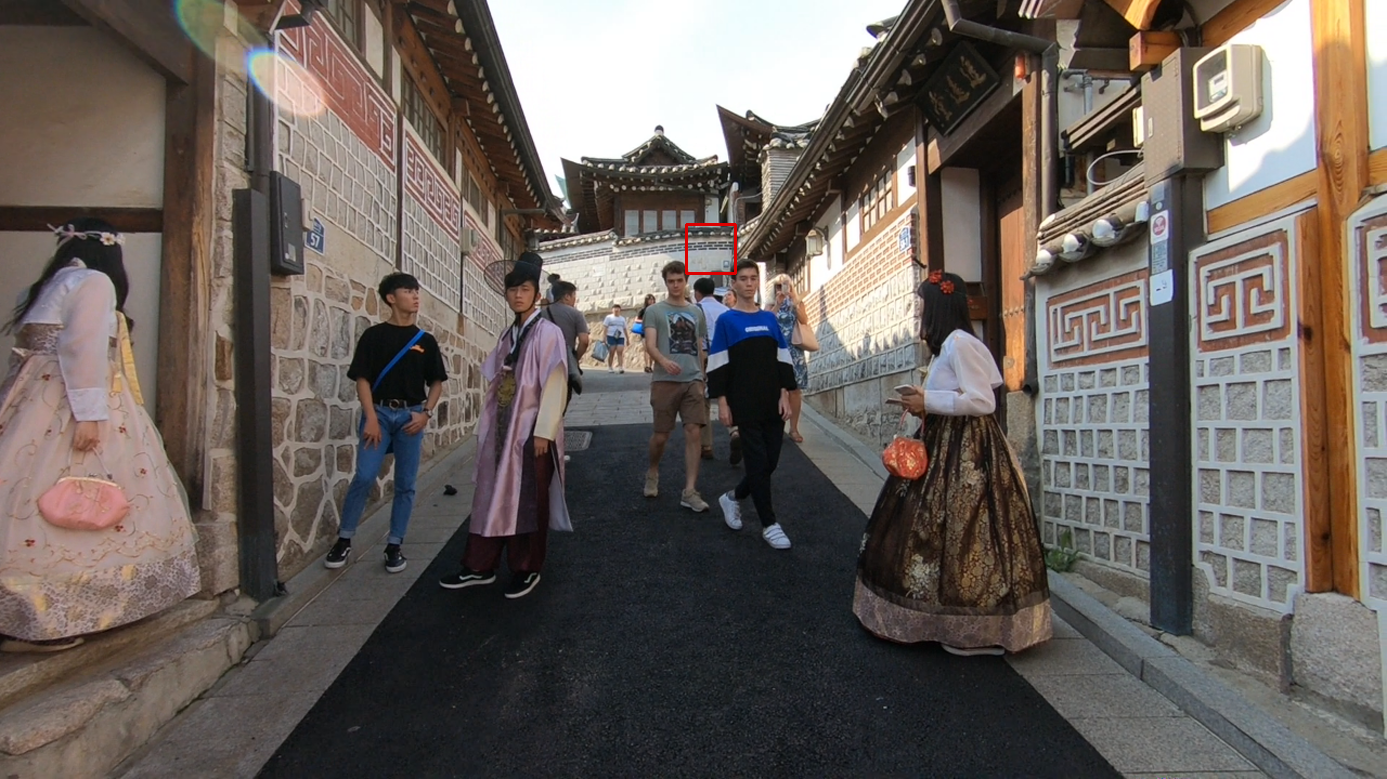}}}
            &  \includegraphics[width=0.15\linewidth]{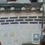}
            &  \includegraphics[width=0.15\linewidth]{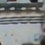}
            &  \includegraphics[width=0.15\linewidth]{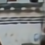}    \\
			\multicolumn{3}{c}{~}                                      &  (a)                 &  (b)                   &  (c)   \\
			\multicolumn{3}{c}{~}
            &  \includegraphics[width=0.15\linewidth]{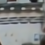}
            & \includegraphics[width=0.15\linewidth]{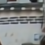}
            & \includegraphics[width=0.15\linewidth]{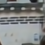}
              \\
			\multicolumn{3}{c}{Frame 092, Clip 011} &   (d)   & (e)      & (f)  \\			
		\end{tabular}
	\end{center}
	\vspace{-6mm}
	\caption{Visual comparisons between Transformer-based methods on the REDS4 dataset. (a) Ground truth. (b)-(f) denote the results generated by VRT~\cite{liang2022vrt}, TTVSR~\cite{liu2022ttvsr}, RVRT~\cite{liang2022rvrt}, PSRT~\cite{shi2022rethinking}, and CFD-PSRT (Ours), respectively. The results in (b)-(d) do not have accurate texture details information. However, PSRT using the proposed CFD (f) can restore much clearer texture details than PSRT (g).}
	\label{fig:visual6}
	\vspace{-6mm}
\end{figure*}


\noindent \textbf{Qualitative comparisons.} We also provide some visual quality comparisons in Figures~\ref{fig:visual3}, ~\ref{fig:visual4}, and~\ref{fig:visual6} to show the effectiveness of our method compared to other approaches.
Figure~\ref{fig:visual3} shows that BasicVSR and BasicVSR++ can not reconstruct the eyes and mouth details without correction method.
Although VRT and TTVSR utilize feature alignment and self-attention mechanism, their restored results have different levels of artifacts.
However, CFD-BasicVSR++ can effectively restore the facial details from video sequence, especially eyes and mouth.
Figure~\ref{fig:visual4} presents that only CFD-BasicVSR and CFD-BasicVSR++ are able to restore the upper-right part of the video frame, while the other methods exhibit various artifacts.
%
Furthermore, Figure~\ref{fig:visual6} shows that our CFD-PSRT can restore much clearer texture details compared to PSRT~\cite{shi2022rethinking}.
More visual comparisons can be found in the supplementary material.

\noindent \textbf{Efficiency comparisons.} By embedding our method into BasciVSR and BasicVSR++, they achieve better performance with only a few parameters increasing.
To achieve a better trade-off between efficiency and performance, we decide not to use any self-attention mechanism in our method due to higher computational complexity and runtime consumption of self-attention.
In comparison to VRT, CFD-BasicVSR++ achieves comparable performance with only 39\% of the runtime consumption.

\begin{table}[t]
\begin{minipage}{1\textwidth}

    \captionof{table}{Quantitative evaluations of the proposed DAC and CFP on the REDS4.}
    \label{tab:ablation}
    \vspace{-6mm}
    \small
    \begin{center}
        \resizebox{0.98\textwidth}{!}{
            \tabcolsep=0.1cm
            \begin{tabular}{l|c|c|c|c|c}
                \toprule
                Methods       & Baseline     & w/ Concat   & w/ DAC \& w/o CFP   & w/o DAC \& w/ CFP  &  w/ DAC \& w/ CFP       \\ \hline
                Parameters (M)    & 6.16            & 6.18                 & 6.16                 & 6.56 & 6.56                \\
                PSNR (dB)         & 31.36               & 31.48                 & 31.57      & 31.50        & \textbf{31.79}      \\
                \bottomrule
            \end{tabular}}
    \end{center}

\footnotesize
\centering
    \begin{tabular}{cccccc}
     \includegraphics[width=0.15\textwidth]{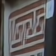}
    &  \includegraphics[width=0.15\textwidth]{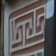}
    & \includegraphics[width=0.15\textwidth]{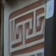} &\includegraphics[width=0.15\textwidth]{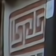}
    & \includegraphics[width=0.15\textwidth]{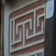}
    &\includegraphics[width=0.15\textwidth]{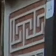} \\
    (a) & (b) & (c) & (d) & (e) & (f)
    \end{tabular}
    \captionof{figure}{Effectiveness of our proposed module for VSR. (a) Baseline. (b)-(e) denote the results by baseline method w/ Concat, w/ DAC \& w/o CFP, w/o DAC \& w/ CFP, and w/ DAC \& w/ CFP, respectively. (f) Ground truth. The methods without using the DAC and CFP contain different artifacts as shown in (b)-(d). In contrast, the proposed approach with the  DAC and CFP generates much clearer structural and edge information in (e).}
    \label{fig:ablation1}
    \vspace{-8mm}
\end{minipage}
\end{table}

\vspace{-3mm}
\section{Analysis and Discussion}
\vspace{-1mm}
\label{sec:analy}
To demonstrate the effectiveness of our proposed method, we further conduct extensive ablation studies and provide additional discussion in this section.
For fair comparison, all the ablation studies are trained on the REDS dataset with 300K iterations and tested on the REDS4 dataset.

Specifically, we compare the proposed method in two ways to show their effectiveness for VSR: 1) we remove all proposed components in CFD-BasicVSR as the first baseline model, and 2) we select BasicVSR~\cite{chan2021basicvsr} as the second baseline model.
Note that the main difference between the first baseline model and BasicVSR is that the first baseline model employs twice forward and backward propagation with fewer propagation blocks, where the parameters are 6.16M and 6.3M, respectively.

\noindent \textbf{Effectiveness of the DAC.~~~~~~~~~~~~~~~~~~~~~~~~~~~~~~~~~~~~~~~~~~~~~~~~~}
\begin{wrapfigure}[9]{r}{0.6\textwidth}
\vspace{-12mm}
    \footnotesize
    \centering
    \begin{tabular}{cccc}
    \includegraphics[width=0.13\textwidth]{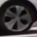}
    & \includegraphics[width=0.13\textwidth]{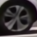}
    & \includegraphics[width=0.13\textwidth]{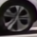}
    & \includegraphics[width=0.13\textwidth]{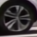} \\
    (a)  & (b)  &(c)  & (d) \\
    \end{tabular}
\vspace{-3mm}
    \captionof{figure}{Effectiveness of our proposed module plugged into BasicVSR~\cite{chan2021basicvsr} for VSR. (a) BasicVSR~\cite{chan2021basicvsr}. (b)-(d) denote the results by BasicVSR~\cite{chan2021basicvsr} w/ DAC,  w/ CFP, and w/ DAC \& w/ CFP, respectively.}
    \label{fig:ablation2}
\end{wrapfigure}
By utilizing structure and edge information of shallow features, the proposed discriminative alignment correction (DAC)~(Eq.\eqref{eq:dac}) compensates for information loss of aligned features during feature warping and effectively reduces the impact of artifacts. We conduct ablation studies for
%
quantitative and qualitative comparisons to demonstrate its effectiveness for VSR.
For quantitative comparisons, Table~\ref{tab:ablation} shows that, compared to baseline model and the proposed method without using the DAC, the baseline using the DAC and the proposed method achieve a PSNR gain of $\sim$0.25dB without any parameter increase.
%
For qualitative comparisons, as shown in Figure~\ref{fig:ablation1}, compared with the baseline model (see Figure~\ref{fig:ablation1}(a)), the baseline model using the DAC (see Figure~\ref{fig:ablation1}(c)) generates much clearer edge details.
Another example in Figure~\ref{fig:ablation2}(b) shows that BasicVSR using the DAC can restore the sharper texture details of the wheel, whereas the counterparts without using the DAC produce blurry outputs (see Figure~\ref{fig:ablation2}(a)).
Furthermore, we replace our DAC with another feature fusion operation, Concatenation, and conduct ablation studies.
Both quantitative and qualitative results in Table~\ref{tab:ablation} and Figure~\ref{fig:ablation1} demonstrate that the method using the DAC achieves better performance improvement and restores much clearer details information without any parameters increasing, compared to the method using concatenation.

\vspace{-8mm}
\noindent \textbf{Effectiveness of the CFP.} The proposed collaborative feedback propagation (CFP) module mainly utilizes different timestep features from both forward and backward propagation for better exploration of spatio-temporal information.
To demonstrate the effectiveness of this module, we further compare the baselines without using the CFP and train these baselines on the same settings for fair comparisons.
Table~\ref{tab:ablation} presents that the proposed method using the CFP generates better results with higher PSNR values and only brings 0.4M parameters gain compared with the proposed method without using the CFP.
%
For qualitative comparisons, Figure~\ref{fig:ablation1}(d) demonstrates that the baseline model using the CFP can effectively model long-range temporal information and produce more accurate edge details.
Moreover, Figure~\ref{fig:ablation2}(a) shows that BasicVSR produces a blurry frame because there is limited information from current features that can be utilized for reconstruction.
%
In contrast, BasicVSR using the CFP (see Figure~\ref{fig:ablation2}(d)) can restore much clearer textures due to the information reinforcement from different timestep features, especially for input frames that contain sharp edges and complex textures.

\vspace{-8mm}
\noindent \textbf{Effectiveness of the GCFB.} We further evaluate the effectiveness of the GCFB in the CFP module.
The GCFB~(\ref{eq:GCFB}) applies the gating mechanism, which brings temporal interaction to the current and future features. One may won-

\hspace{-6mm}
\begin{minipage}{\textwidth}
\begin{minipage}{0.38\textwidth}
der whether existing methods based on gating mechanisms, \eg, GDFN~\cite{Zamir2021Restormer} performs better or not.
To answer this question, we conduct ablation studies by replacing GCFBs with Resblocks~\cite{he2016deep} (w/o gating), GDFNs~\cite{Zamir2021Restormer}, and GDFNs with temporal interaction, to verify the effect of GCFBs.
The $2^{th}$ column of Table~\ref{tab:GCFB} shows that the CFP with Resblocks leads to performance degradation.~ To ~evaluate~ the
\end{minipage}
\hspace{0.5mm}
\begin{minipage}{0.6\textwidth}
\vspace{-4mm}
    \captionof{table}{Effect of the proposed HU on the Vid4.}
    \centering
    \footnotesize
    \resizebox{\textwidth}{!}{
    \begin{tabular}{l|c|c|c|c}
        \toprule
            Methods             & Resblocks~\cite{he2016deep}      & GDFNs~\cite{Zamir2021Restormer}     & GDFNs w/ Tem.    & GCFBs        \\ \hline
            Paras.(M) & 6.81 & 6.56 & 6.57 & 6.56 \\
                PSNR (dB)          & 31.69  & 31.63   & 31.70  & \textbf{31.79}          \\
                \bottomrule
    \end{tabular}
    }
    \label{tab:GCFB}
    \footnotesize
    \centering
    \begin{tabular}{cccc}
     \includegraphics[width=0.23\textwidth]{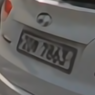}
    & \includegraphics[width=0.23\textwidth]{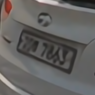}
    &  \includegraphics[width=0.23\textwidth]{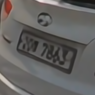}
    &  \includegraphics[width=0.23\textwidth]{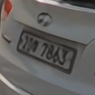} \\
    (a)  & (b) & (c)  & (d) \\
    \end{tabular}
    \vspace{-1mm}
    \captionof{figure}{Effectiveness of our proposed GCFB for VSR. (a)-(d) denote the results by proposed methods using Resblocks~\cite{he2016deep}, GDFNs~\cite{Zamir2021Restormer}, GDFNs with temporal interaction, and GCFBs, respectively.}
    \label{fig:ablation3}
\end{minipage}
\end{minipage}

\vspace{1.5mm}
\noindent effectiveness of GCFB compared to GDFN in VSR tasks, we conduct two settings for GDFN: the first setting is the original GDFN using $r_t$ as input; and the second setting concatenates $h_{t+1}$ and $r_t$ as input, bringing temporal interaction to GDFN.
%
whether existing methods based on gating mechanisms, \eg, GDFN~\cite{Zamir2021Restormer} performs better or not.
To answer this question, we conduct ablation studies by replacing GCFBs with Resblocks~\cite{he2016deep} (w/o gating), GDFNs~\cite{Zamir2021Restormer}, and GDFNs with temporal interaction, to verify the effect of GCFBs.
The $2^{th}$ column of Table~\ref{tab:GCFB} shows that the CFP with Resblocks leads to performance degradation.
%
To evaluate the effectiveness of GCFB compared to GDFN in VSR tasks, we conduct two settings for GDFN: the first setting is the original GDFN using $r_t$ as input; and the second setting concatenates $h_{t+1}$ and $r_t$ as input, bringing temporal interaction to GDFN.
Compared with the original GDFNs, GCFBs achieve better quantitative results, indicating that the method only depending on the spatial information is not suitable in VSR.
Introducing temporal interactions to GDFNs leads to slightly improvement, but it still does not perform well compared to the GCFBs.
Figure~\ref{fig:ablation3} further demonstrates that using our proposed GCFBs generates much clearer characters of the license plate.

\section{Conclusion}
In this paper, we have presented an effective collaborative feedback discriminative propagation method for VSR.
We develop a discriminative alignment correction method to reduce the influences of artifacts caused by inaccurate alignment.
To better explore spatio-temporal information during the propagation process and reduce errors accumulated by inaccurate aligned features, we propose a collaborative feedback propagation module, which simultaneously leverages spatio-temporal information of different timestep features from forward and backward propagation.
To demonstrate the effectiveness of our method, we conduct extensive experiments and analysis by integrating it into existing VSR backbones.
Both quantitative and qualitative results show that the proposed method performs favorably against state-of-the-art methods on several benchmark datasets.

%
%
\bibliographystyle{splncs04}
\bibliography{reference}

\begin{thebibliography}{10}
\providecommand{\url}[1]{\texttt{#1}}
\providecommand{\urlprefix}{URL }
\providecommand{\doi}[1]{https://doi.org/#1}

\bibitem{ba2016layer}
Ba, J.L., Kiros, J.R., Hinton, G.E.: Layer normalization. arXiv preprint
  arXiv:1607.06450  (2016)

\bibitem{cao2021vsrt}
Cao, J., Li, Y., Zhang, K., Van~Gool, L.: Video super-resolution transformer.
  arXiv preprint arXiv:2106.06847  (2021)

\bibitem{chan2021basicvsr}
Chan, K.C., Wang, X., Yu, K., Dong, C., Loy, C.C.: {BasicVSR}: The search for
  essential components in video super-resolution and beyond. In: CVPR (2021)

\bibitem{chan2022basicvsrplusplus}
Chan, K.C., Zhou, S., Xu, X., Loy, C.C.: {BasicVSR++}: Improving video
  super-resolution with enhanced propagation and alignment. In: CVPR (2022)

\bibitem{charbonnier1994two}
Charbonnier, P., Blanc-Feraud, L., Aubert, G., Barlaud, M.: Two deterministic
  half-quadratic regularization algorithms for computed imaging. In: ICIP
  (1994)

\bibitem{chen2021robust}
Chen, C., Li, H.: Robust representation learning with feedback for single image
  deraining. In: CVPR (2021)

\bibitem{cho2021rethinking}
Cho, S.J., Ji, S.W., Hong, J.P., Jung, S.W., Ko, S.J.: Rethinking
  coarse-to-fine approach in single image deblurring. In: ICCV (2021)

\bibitem{deng2021feedback}
Deng, X., Zhang, Y., Xu, M., Gu, S., Duan, Y.: Deep coupled feedback network
  for joint exposure fusion and image super-resolution. IEEE TIP pp. 3098--3112
  (2021)

\bibitem{dong2014learning}
Dong, C., Loy, C.C., He, K., Tang, X.: Learning a deep convolutional network
  for image super-resolution. In: ECCV (2014)

\bibitem{haris2019recurrent}
Haris, M., Shakhnarovich, G., Ukita, N.: Recurrent back-projection network for
  video super-resolution. In: CVPR (2019)

\bibitem{haris2018deep}
Haris, M., Shakhnarovich, G., Ukita, N.: Deep back-projection networks for
  super-resolution. In: CVPR (2018)

\bibitem{he2016deep}
He, K., Zhang, X., Ren, S., Sun, J.: Deep residual learning for image
  recognition. In: CVPR (2016)

\bibitem{isobe2020video}
Isobe, T., Jia, X., Gu, S., Li, S., Wang, S., Tian, Q.: Video super-resolution
  with recurrent structure-detail network. In: ECCV (2020)

\bibitem{isobe2022look}
Isobe, T., Jia, X., Tao, X., Li, C., Li, R., Shi, Y., Mu, J., Lu, H., Tai,
  Y.W.: Look back and forth: video super-resolution with explicit temporal
  difference modeling. In: CVPR (2022)

\bibitem{jo2018deep}
Jo, Y., Wug~Oh, S., Kang, J., Joo~Kim, S.: Deep video super-resolution network
  using dynamic upsampling filters without explicit motion compensation. In:
  CVPR (2018)

\bibitem{kim2018spatio}
Kim, T.H., Sajjadi, M.S.M., Hirsch, M., Sch\"{o}lkopf, B.: Spatio-temporal
  transformer network for video restoration. In: ECCV (2018)

\bibitem{kingma2014adam}
Kingma, D., Ba, J.: Adam: A method for stochastic optimization. In: ICLR (2015)

\bibitem{li2019gated}
Li, Q., Li, Z., Lu, L., Jeon, G., Liu, K., Yang, X.: Gated multiple feedback
  network for image super-resolution. In: BMVC (2019)

\bibitem{li2020mucan}
Li, W., Tao, X., Guo, T., Qi, L., Lu, J., Jia, J.: {MuCAN}:
  Multi-correspondence aggregation network for video super-resolution. In: ECCV
  (2020)

\bibitem{li2019feedback}
Li, Z., Yang, J., Liu, Z., Yang, X., Jeon, G., Wu, W.: Feedback network for
  image super-resolution. In: CVPR (2019)

\bibitem{liang2022vrt}
Liang, J., Cao, J., Fan, Y., Zhang, K., Ranjan, R., Li, Y., Timofte, R.,
  Van~Gool, L.: {VRT}: A video restoration transformer. arXiv preprint
  arXiv:2201.12288  (2022)

\bibitem{liang2022rvrt}
Liang, J., Fan, Y., Xiang, X., Ranjan, R., Ilg, E., Green, S., Cao, J., Zhang,
  K., Timofte, R., Van~Gool, L.: Recurrent video restoration transformer with
  guided deformable attention. In: NIPS (2022)

\bibitem{seq2seq}
Lin, J., Hu, X., Cai, Y., Wang, H., Yan, Y., Zou, X., Zhang, Y., Van~Gool, L.:
  Unsupervised flow-aligned sequence-to-sequence learning for video
  restoration. In: ICML (2022)

\bibitem{liu2014bayesian}
Liu, C., Sun, D.: On bayesian adaptive video super resolution. IEEE TPAMI pp.
  346--360 (2014)

\bibitem{liu2022ttvsr}
Liu, C., Yang, H., Fu, J., Qian, X.: Learning trajectory-aware transformer for
  video super-resolution. In: CVPR (2022)

\bibitem{loshchilov2016sgdr}
Loshchilov, I., Hutter, F.: {SGDR}: Stochastic gradient descent with warm
  restarts. arXiv preprint arXiv:1608.03983  (2016)

\bibitem{nah2019ntire}
Nah, S., Baik, S., Hong, S., Moon, G., Son, S., Timofte, R., Mu~Lee, K.:
  {NTIRE} 2019 challenge on video deblurring and super-resolution: Dataset and
  study. In: CVPRW (2019)

\bibitem{Pan_2023_CVPR}
Pan, J., Xu, B., Dong, J., Ge, J., Tang, J.: Deep discriminative spatial and
  temporal network for efficient video deblurring. In: CVPR (2023)

\bibitem{ranjan2017optical}
Ranjan, A., Black, M.J.: Optical flow estimation using a spatial pyramid
  network. In: CVPR (2017)

\bibitem{sajjadi2018frame}
Sajjadi, M.S.M., Vemulapalli, R., Brown, M.: Frame-recurrent video
  super-resolution. In: CVPR (2018)

\bibitem{shi2022rethinking}
Shi, S., Gu, J., Xie, L., Wang, X., Yang, Y., Dong, C.: Rethinking alignment in
  video super-resolution transformers. In: NIPS (2022)

\bibitem{shi2016real}
Shi, W., Caballero, J., Husz{\'a}r, F., Totz, J., Aitken, A.P., Bishop, R.,
  Rueckert, D., Wang, Z.: Real-time single image and video super-resolution
  using an efficient sub-pixel convolutional neural network. In: CVPR (2016)

\bibitem{Sun2018PWC-Net}
Sun, D., Yang, X., Liu, M.Y., Kautz, J.: {PWC-Net}: {CNNs} for optical flow
  using pyramid, warping, and cost volume. In: CVPR (2018)

\bibitem{tao2017detail}
Tao, X., Gao, H., Liao, R., Wang, J., Jia, J.: Detail-revealing deep video
  super-resolution. In: CVPR (2017)

\bibitem{tian2020tdan}
Tian, Y., Zhang, Y., Fu, Y., Xu, C.: {TDAN}: Temporally deformable alignment
  network for video super-resolution. In: CVPR (2018)

\bibitem{wang2018video}
Wang, T.C., Liu, M.Y., Zhu, J.Y., Liu, G., Tao, A., Kautz, J., Catanzaro, B.:
  Video-to-video synthesis. arXiv preprint arXiv:1808.06601  (2018)

\bibitem{wang2019edvr}
Wang, X., Chan, K.C., Yu, K., Dong, C., Loy, C.C.: {EDVR}: Video restoration
  with enhanced deformable convolutional networks. In: CVPRW (2019)

\bibitem{xiang2020zooming}
Xiang, X., Tian, Y., Zhang, Y., Fu, Y., Allebach, J.P., Xu, C.: Zooming
  slow-mo: Fast and accurate one-stage space-time video super-resolution. In:
  CVPR (2020)

\bibitem{xue2019video}
Xue, T., Chen, B., Wu, J., Wei, D., Freeman, W.T.: Video enhancement with
  task-oriented flow. IJCV pp. 1106--1125 (2019)

\bibitem{yi2019progressive}
Yi, P., Wang, Z., Jiang, K., Jiang, J., Ma, J.: Progressive fusion video
  super-resolution network via exploiting non-local spatio-temporal
  correlations. In: ICCV (2019)

\bibitem{ying2020deformable}
Ying, X., Wang, L., Wang, Y., Sheng, W., An, W., Guo, Y.: Deformable 3d
  convolution for video super-resolution. IEEE SPL pp. 1500--1504 (2020)

\bibitem{zamir2017feedback}
Zamir, A.R., Wu, T.L., Sun, L., Shen, W.B., Shi, B.E., Malik, J., Savarese, S.:
  Feedback networks. In: CVPR (2017)

\bibitem{Zamir2021Restormer}
Zamir, S.W., Arora, A., Khan, S., Hayat, M., Khan, F.S., Yang, M.H.: Restormer:
  Efficient transformer for high-resolution image restoration. In: CVPR (2022)

\bibitem{zhang2018image}
Zhang, Y., Li, K., Li, K., Wang, L., Zhong, B., Fu, Y.: Image super-resolution
  using very deep residual channel attention networks. In: ECCV (2018)

\end{thebibliography}


\begin{thebibliography}{10}
\providecommand{\url}[1]{\texttt{#1}}
\providecommand{\urlprefix}{URL }
\providecommand{\doi}[1]{https://doi.org/#1}

\bibitem{chan2021basicvsr}
Chan, K.C., Wang, X., Yu, K., Dong, C., Loy, C.C.: {BasicVSR}: The search for
  essential components in video super-resolution and beyond. In: CVPR (2021)

\bibitem{chan2022basicvsrplusplus}
Chan, K.C., Zhou, S., Xu, X., Loy, C.C.: {BasicVSR++}: Improving video
  super-resolution with enhanced propagation and alignment. In: CVPR (2022)

\bibitem{chan2022investigating}
Chan, K.C., Zhou, S., Xu, X., Loy, C.C.: Investigating tradeoffs in real-world
  video super-resolution. In: CVPR (2022)

\bibitem{charbonnier1994two}
Charbonnier, P., Blanc-Feraud, L., Aubert, G., Barlaud, M.: Two deterministic
  half-quadratic regularization algorithms for computed imaging. In: ICIP
  (1994)

\bibitem{cho2021rethinking}
Cho, S.J., Ji, S.W., Hong, J.P., Jung, S.W., Ko, S.J.: Rethinking
  coarse-to-fine approach in single image deblurring. In: ICCV (2021)

\bibitem{he2016deep}
He, K., Zhang, X., Ren, S., Sun, J.: Deep residual learning for image
  recognition. In: CVPR (2016)

\bibitem{isobe2020video}
Isobe, T., Jia, X., Gu, S., Li, S., Wang, S., Tian, Q.: Video super-resolution
  with recurrent structure-detail network. In: ECCV (2020)

\bibitem{liang2022vrt}
Liang, J., Cao, J., Fan, Y., Zhang, K., Ranjan, R., Li, Y., Timofte, R.,
  Van~Gool, L.: {VRT}: A video restoration transformer. arXiv preprint
  arXiv:2201.12288  (2022)

\bibitem{liu2014bayesian}
Liu, C., Sun, D.: On bayesian adaptive video super resolution. IEEE TPAMI pp.
  346--360 (2014)

\bibitem{liu2022ttvsr}
Liu, C., Yang, H., Fu, J., Qian, X.: Learning trajectory-aware transformer for
  video super-resolution. In: CVPR (2022)

\bibitem{nah2019ntire}
Nah, S., Baik, S., Hong, S., Moon, G., Son, S., Timofte, R., Mu~Lee, K.:
  {NTIRE} 2019 challenge on video deblurring and super-resolution: Dataset and
  study. In: CVPRW (2019)

\bibitem{wang2019edvr}
Wang, X., Chan, K.C., Yu, K., Dong, C., Loy, C.C.: {EDVR}: Video restoration
  with enhanced deformable convolutional networks. In: CVPRW (2019)

\bibitem{xue2019video}
Xue, T., Chen, B., Wu, J., Wei, D., Freeman, W.T.: Video enhancement with
  task-oriented flow. IJCV pp. 1106--1125 (2019)

\bibitem{yang2021real}
Yang, X., Xiang, W., Zeng, H., Zhang, L.: Real-world video super-resolution: A
  benchmark dataset and a decomposition based learning scheme. In: ICCV (2021)

\bibitem{yi2019progressive}
Yi, P., Wang, Z., Jiang, K., Jiang, J., Ma, J.: Progressive fusion video
  super-resolution network via exploiting non-local spatio-temporal
  correlations. In: ICCV (2019)

\end{thebibliography}
\end{document}


\title{Collaborative Feedback Discriminative Propagation for Video Super-Resolution\\ - Supplemental Material -}

\titlerunning{Feedback Discriminative Propagation for Video Super-Resolution}

\author{Hao Li \and
Xiang Chen \and
Jiangxin Dong \and
Jinhui Tang \and
Jinshan Pan}

\authorrunning{H.~Li et al.}

\institute{Nanjing University of Science and Technology}

\maketitle

\section*{\centering Overview}
%
In this document, we first provide more visual comparisons of the aligned features with our discriminative alignment correction method in Section~\ref{sec:dac}.
Then, we describe more details on datasets and experimental settings of our models in Section~\ref{sec:details}.
Moreover, we evaluate the temporal consistency by comparing the temporal profiles between our models and some commonly-used video super-resolution (VSR) backbones in Section~\ref{sec:tem}.
Furthermore, we discuss the limitation of our method in Section~\ref{sec:limit}.
Finally, we show more qualitative results compared to state-of-the-art methods in Section~\ref{sec:visual}.

\begin{figure*}[ht]
\tiny
\centering
    \begin{tabular}{ccccc}
     \includegraphics[width=0.19\textwidth]{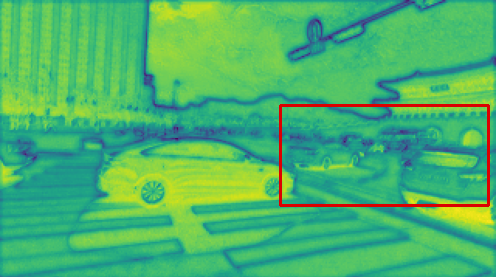}  & \includegraphics[width=0.19\textwidth]{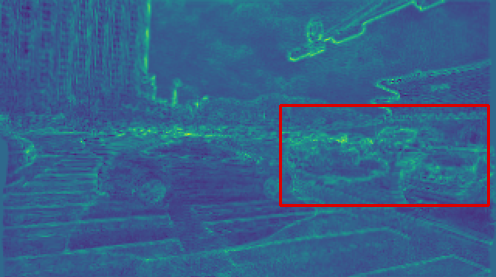}  & \includegraphics[width=0.19\textwidth]{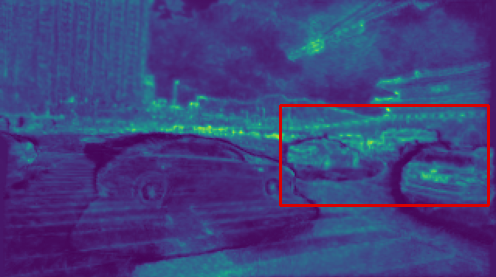}  & \includegraphics[width=0.19\textwidth]{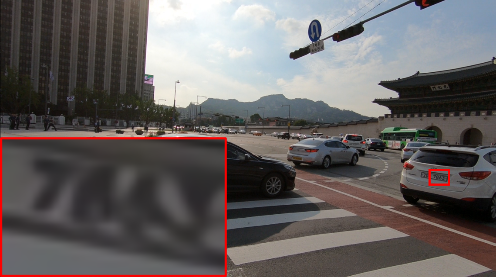}  & \includegraphics[width=0.19\textwidth]{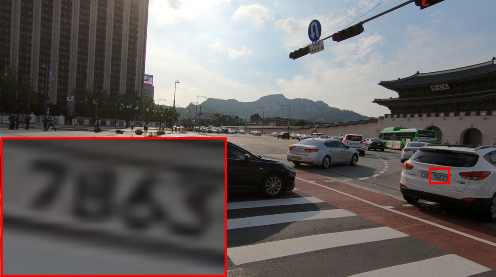} \\
     \includegraphics[width=0.19\textwidth]{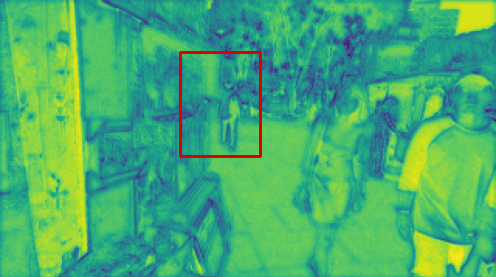}  & \includegraphics[width=0.19\textwidth]{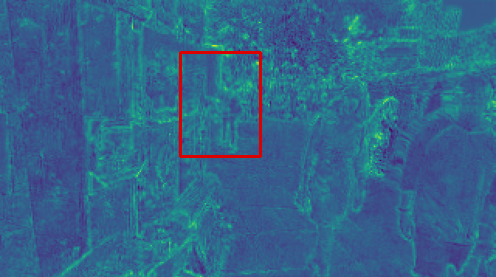}  & \includegraphics[width=0.19\textwidth]{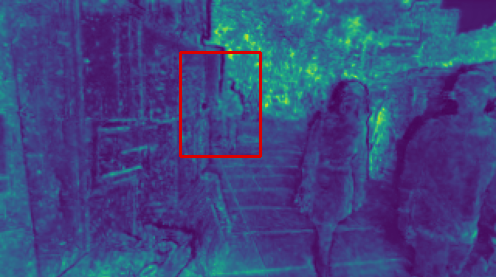}  & \includegraphics[width=0.19\textwidth]{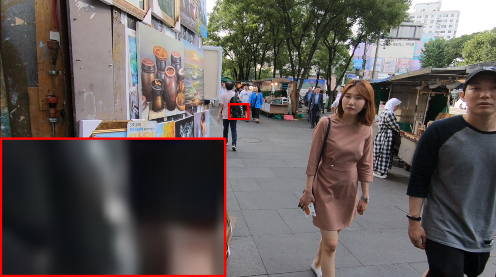}  & \includegraphics[width=0.19\textwidth]{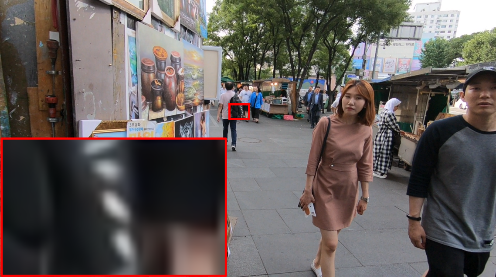} \\
     \includegraphics[width=0.19\textwidth]{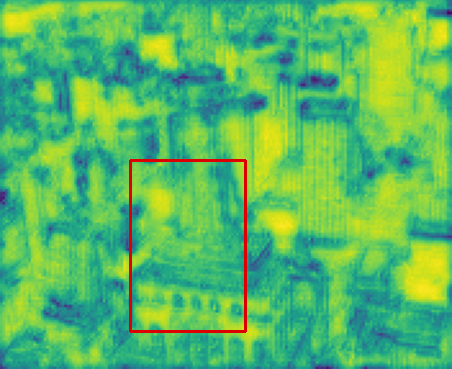}  & \includegraphics[width=0.19\textwidth]{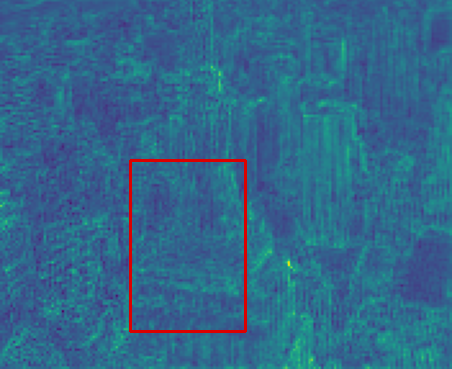}  & \includegraphics[width=0.19\textwidth]{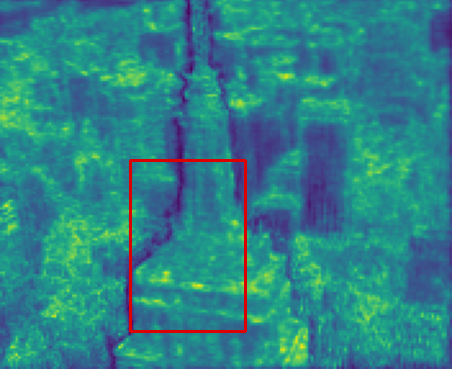}  & \includegraphics[width=0.19\textwidth]{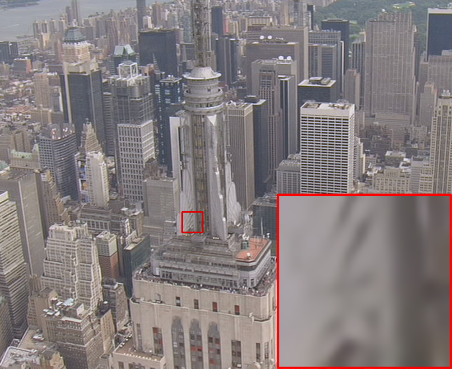}  & \includegraphics[width=0.19\textwidth]{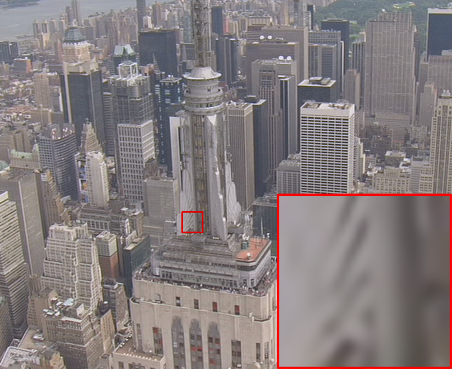} \\
     (a) Shallow features  & (b) Feature alignment  & (c) DAC  & (d) Results with (b)  & (e) Results with (c)
    \end{tabular}
    \caption{\textbf{Effect of the proposed DAC method on VSR.} The aligned features with feature alignment have severe artifacts, leading to regions with significant structures and details having a \textit{lower contrast}. In contrast, the shallow features (b) contain better structures and details. By using our proposed DAC method in combination with shallow features, the contrast in these regions becomes \textit{higher} (see \rf{red} box), and the artifacts can be effectively corrected. Moreover, the restored VSR results (e) have better visual quality than (d).}
    \label{fig:dac}
\vspace{-8mm}
\end{figure*}

\section{More Visual Comparisons with Our DAC} \label{sec:dac}
To further demonstrate the effectiveness of our proposed discriminative alignment correction (DAC) method, we provide more visual comparisons between the aligned features with DAC and those with feature alignment, as shown in Figure~\ref{fig:dac}.
As we claimed in the main paper, the artifacts in aligned features usually destroy the main structures and details, which leads to the features with significant structures or details having a \textit{lower contrast} (see \rf{red} box in Figure~\ref{fig:dac}(b)).
However, our DAC method can utilize the information from shallow features to discriminatively select the regions with a higher response, resulting in these regions having \textit{higher contrast} (see \rf{red} box in Figure~\ref{fig:dac}(c)).
Figure~\ref{fig:dac}(e) shows that, compared to the VSR result with feature alignment, our DAC can restore clearer structures and details.

\section{Datasets, Experimental Settings and Results} \label{sec:details}
\noindent \textbf{Training and testing datasets.} We train the models on two different training datasets (\textit{i.e.,} REDS~\cite{nah2019ntire} and Vimeo-90K~\cite{xue2019video}) for $4\times$ VSR.
We utilize the MATLAB $\mathrm{imresize}$ function to generate low-resolution (LR) images corresponding to REDS (Bicubic degradation) and Vimeo-90K (Bicubic degradation, and blur downsampling with $\sigma=1.6$ Gaussian blur).
For REDS training, we use REDS4~\cite{wang2019edvr} as the test dataset, while Vimeo-T~\cite{xue2019video}, Vid4~\cite{liu2014bayesian} and UDM10~\cite{yi2019progressive} are used as the test datasets for Vimeo-90K training.
On a server with 8 NVIDIA GeForce RTX 3090 GPUs, it takes about 4, 10, and 28 days to train our CFD-BasicVSR, CFD-BasicVSR++, and CFD-PSRT, respectively.

\noindent \textbf{REDS.} REDS~\cite{nah2019ntire} is a high-quality video dataset for VSR, which has 270 clips (each clip has 100 frames) with a size of $1280 \times 720$ for training and testing.
We follow~\cite{wang2019edvr} to select 4 clips (\textit{i.e.,} Clip 000, 011, 015, 020) as REDS4 benchmark for testing and the rest 266 clips for training.

\noindent \textbf{Vimeo-90K.} Vimeo-90K~\cite{xue2019video} is a middle-quality video dataset that contains 72,436 clips (each clip has 7 frames) with a size of $448 \times 256$ for VSR.
For the VSR task, It uses 64,612 clips for training and the rest 7,824 clips for testing (denoted as Vimeo-T).

\noindent \textbf{Vid4.} Vid4~\cite{liu2014bayesian} is a testing dataset for VSR, which consists of 4 video clips (\textit{i.e.,} calendar, city, foliage and walk), and each clip has about 40 frames with a size of $720 \times 480$.

\noindent \textbf{UDM10.} UDM10~\cite{yi2019progressive} is a newly-proposed testing dataset for VSR that contains 10 clips, each of which has 32 frames with a size of $1272 \times 720$.

\begin{table}[!ht]
    \caption{The number of residual blocks~\cite{he2016deep} used in our CNN-based models.}
    \centering

     \resizebox{0.8\textwidth}{!}{
            \begin{tabular}{l|c|c}
            \toprule
                Methods                              & ~~CFD-BasicVSR~~ & ~CFD-BasicVSR++~  \\ \hline
                Feature Extractor                    & 5            & 5 \\
                Forward/Backward Prop. Module        & 10           & 7 \\
                Reconstruction Module                & 3            & 5 \\
            \bottomrule
            \end{tabular}
    }

    \label{tab:resblock}
    \vspace{-4mm}
\end{table}

\begin{table}[t]
    \centering
    \caption{The number of input frames and training iterations for our models.}
    \vspace{-4mm}
    \begin{center}
        \resizebox{0.8\textwidth}{!}{
            \begin{tabular}{l||c|c||c|c||c}
            \toprule
                \multirow{2}{*}{Methods}             & \multicolumn{2}{c||}{CFD-BasicVSR}    & \multicolumn{2}{c||}{CFD-BasicVSR++}  & CFD-PSRT   \\
                \cline{2-6}                          & ~~REDS~~           & ~Vimeo-90K~            & ~~REDS~~        & ~Vimeo-90K~   & ~~REDS~~  \\ \hline
                Number of Frames~~~                    & 15             & 7                    & 30           & 7     & 30               \\
                Total Iterations                     & 300,000        & 300,000              & 600,000      & 300,000     &600,000         \\
            \bottomrule
            \end{tabular}
        }
    \end{center}
    \label{tab:settings}
    \vspace{-4mm}
\end{table}

\begin{table}[t]
    \caption{Ablation study on the numbers of the proposed GCFB.}
    \vspace{-4mm}
    \label{tab:GCFB}
    \begin{center}
            \resizebox{0.8\textwidth}{!}{
            \begin{tabular}{l|c|c|c|c|c}
                \toprule
                Numbers                    & 1     & 3      & 5               & 7      & Residual blocks (5) \\ \hline
                Parameters (M)             & 6.46  & 6.51   & 6.56            & 6.62   & 6.81            \\
                FLOPs (G)                  & 137   & 139    & 141             & 142    & 150             \\
                PSNR (dB)                  & 31.67 & 31.74  & \textbf{31.79}  & 31.77  & 31.69           \\
                \bottomrule
            \end{tabular}}
    \end{center}
    \vspace{-4mm}
\end{table}

\noindent \textbf{Parameters settings.} We design three new VSR models, CFD-BasicVSR, CFD-BasicVSR++, and CFD-PSRT respectively.
As shown in Figure 2 of the main paper, our model contains a feature extractor, a forward/backward propagation module with the discriminative alignment correction (DAC), a collaborative feedback propagation (CFP) module, and a reconstruction module.
%
Table~\ref{tab:resblock} shows the settings for the number of residual blocks used in each component of our CNN-based models.
During REDS training, CFD-BasicVSR uses 15 LR frames as inputs, while CFD-BasicVSR++ and CFD-PSRT uses 30 LR frames.
Since each sequence in Vimeo-90K has only 7 frames, both our models use 7 LR frames as inputs for Vimeo-90K training.
In particular, CFD-BasicVSR++ uses the pre-trained model on REDS to initialize the model parameters when trained on Vimeo-90K.
The total number of training iterations is presented in Table~\ref{tab:settings}.

\noindent \textbf{Loss functions.} We utilize Charbonnier~\cite{charbonnier1994two} and FFT~\cite{cho2021rethinking} loss to constrain the network training.
The Charbonnier loss is defined as follows:
\begin{equation}
    \mathcal{L}_{char} = \sqrt{\|I_{SR}-I_{GT}\|^{2}+\varepsilon^{2}},
    \label{equ:closs}
\end{equation}
where $I_{SR}$ and $I_{GT}$ denote the network output and the corresponding grouth truth frame, and $\varepsilon$ is an empirical value.
However, as demonstrated by~\cite{cho2021rethinking}, only using the pixel-wise loss function (\textit{i.e.,} Charbonnier loss) does not effectively help the detail estimation.
Thus, we further employ the constraint from the frequency domain to regularize network training.
The total loss function for the network training is defined as follows:
\begin{equation}
    \mathcal{L} = \mathcal{L}_{char} + \lambda \|\mathcal{F}(I_{SR})-\mathcal{F}(I_{GT}) \|_{1},
    \label{equ:loss}
\end{equation}
where $\mathcal{F}$ denotes the fast Fourier transform, and we set $\lambda=0.1$ in all experiments.

\noindent \textbf{The numbers of GCFBs.} As shown in Table~\ref{tab:GCFB}, we conduct ablation studies to verify the impact of different numbers of GCFBs.
With increasing in the number of GCFBs from 1 to 5, we achieve an improvement of 0.12dB in PSNR index, and corresponding increases in parameters and Flops.
However, when we set the number of GCFBs to 7, the performance drop by 0.02dB in PSNR index.
Considering the trade-off between efficiency and performance, we set the numbers of GCFBs to 5 as the default setting.

\noindent \textbf{Comparisons of FLOPs.} We provide the comparisons of FLOPs with other SOTA VSR methods. As shown in Table~\ref{table:flops}, our methods achieve a better trade-off between efficiency and performance.

\begin{table}[t]
	\centering
    \small
	\caption{Quantitative comparison on the REDS4. FLOPs are caculated on three $180 \times 320$ frames.}
	\vspace{-7mm}
	\begin{center}
    \resizebox{\textwidth}{!}{
	\begin{tabular}{cccccccc}
	\hline
	Methods        & EDVR    & BasicVSR & BasicVSR++ & VRT     & TTVSR   & CFD-BasicVSR    &  CFD-BasicVSR++  \\ \hline
	FLOPs (G)      & 1548.8  & \textbf{1094.8}   & 1121.5     & 3930.2  & 1963.8  & 1185.3  & 1253.1  \\
	Parameters (M) & 20.6    & \textbf{6.3}      & 7.3        & 35.6    & 6.8     & 6.6     & 7.5     \\
    PSNR (dB)      & 31.09   & 31.42    & 32.39      & 32.19   & 32.12   & 31.79   & \textbf{32.51}   \\
 \hline
    \end{tabular}
	}
    \end{center}
	\label{table:flops}
	\vspace{-4mm}
\end{table}

\begin{figure*}[t]
	\tiny
	\begin{center}
		\begin{tabular}{c c c c c c c c c}
			\multicolumn{3}{c}{\multirow{2}*[68pt]{\hspace{-8mm} \includegraphics[width=0.5\linewidth,height=0.4\linewidth]{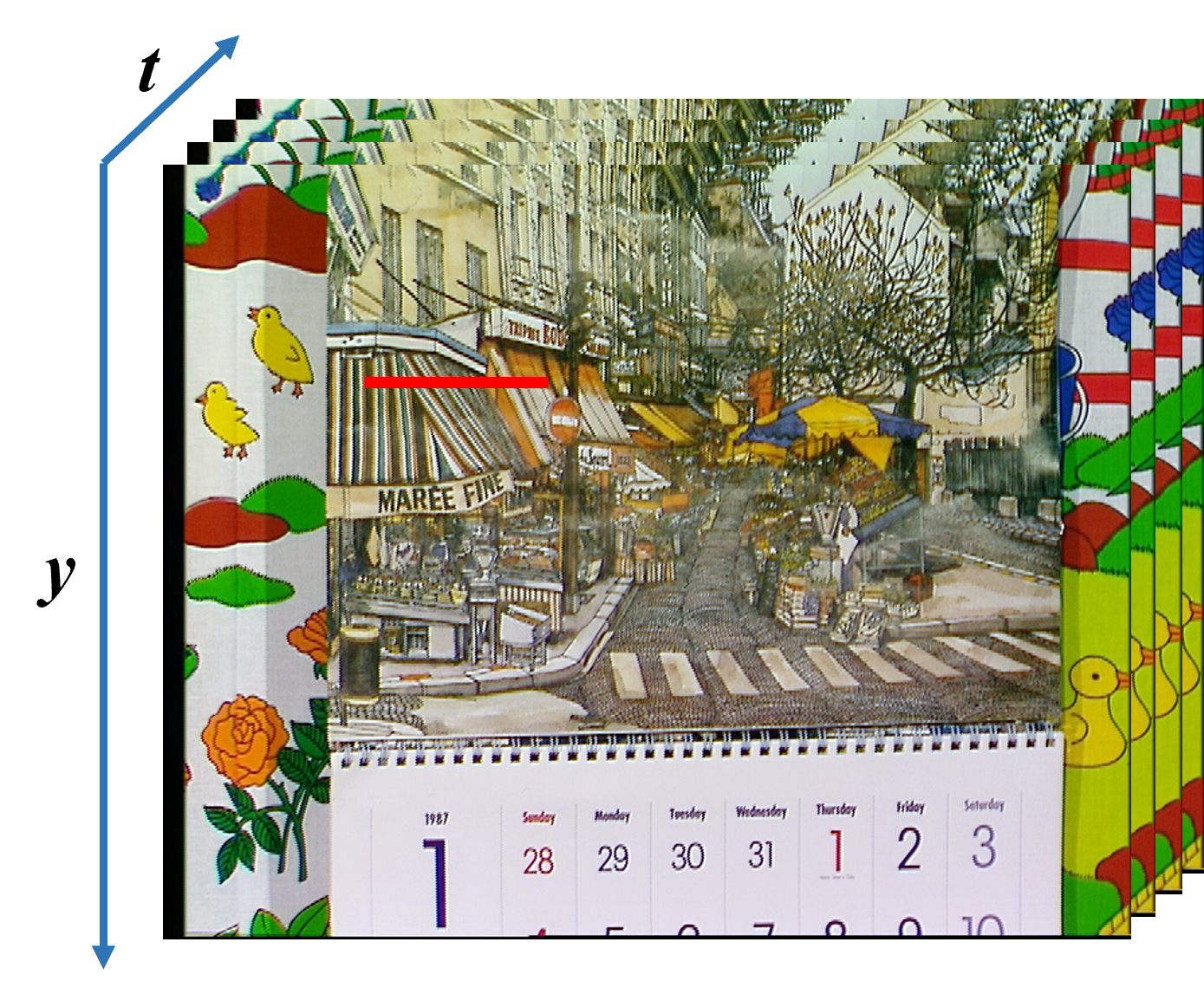}}}
            & \includegraphics[width=0.12\linewidth,height=0.18\linewidth]{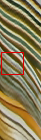}  & \rotatebox{90}{{~~~~~~~~~EDVR}}
            & \includegraphics[width=0.12\linewidth,height=0.18\linewidth]{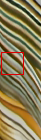}  &  \rotatebox{90}{~~~~~~BasicVSR}
            &  \includegraphics[width=0.12\linewidth,height=0.18\linewidth]{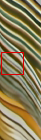}  &  \rotatebox{90}{~~~~~BasicVSR++}      \\
			\multicolumn{3}{c}{~}
            & \includegraphics[width=0.12\linewidth,height=0.18\linewidth]{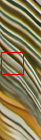} &  \rotatebox{90}{~~~~~~~~~~VRT}
            &  \includegraphics[width=0.12\linewidth,height=0.18\linewidth]{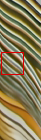} &  \rotatebox{90}{~CFD-BasicVSR++}
            &  \includegraphics[width=0.12\linewidth,height=0.18\linewidth]{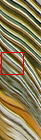} &  \rotatebox{90}{~~~~~~~~~~~GT} \\
		\end{tabular}
	\end{center}
	\vspace{-2mm}
	\caption{Temporal profile visualization for the \rf{red} line on the \textit{calendar} clip~\cite{liu2014bayesian} to show the temporal consistency. The profile from EDVR~\cite{wang2019edvr} has noise, indicating artifacts in the output video. And the profiles from BasicVSR~\cite{chan2021basicvsr}, BasicVSR++~\cite{chan2022basicvsrplusplus}, and VRT~\cite{liang2022vrt} show different level of discontinuity. Benefit from our DAC and CFP method, CFD-BasicVSR++ achieves the most consistent result with smoother temporal transition. Zooming in for better views.}
	\label{fig:temporal}
	\vspace{-2mm}
\end{figure*}


\section{Temporal Consistency} \label{sec:tem}
Following~\cite{isobe2020video,yi2019progressive,chan2022basicvsrplusplus}, we use the temporal profile to qualitatively compare temporal consistency.
As shown in Figure~\ref{fig:temporal}, we take horizontal rows of pixels from the same location (see \rf{red} line in Figure~\ref{fig:temporal}) in consecutive frames and stack them vertically to generate temporal profiles from the \textit{calendar} clip from Vid4~\cite{liu2014bayesian}.
Compared to state-of-the-art methods, our CFD-BasicVSR++ achieves the most consistent result with smoother temporal transition, while BasicVSR and VRT produce discontinuous profiles.
%
%
%
We include more video results in our project website \url{https://github.com/House-Leo/CFDVSR}.

\section{Limitations} \label{sec:limit}
Similar to other recurrent frameworks, our method still requires caching the intermediate features during propagation, \textit{i.e.,} the memory footprint would increase with the length of the input sequence.
Moreover, the depth-wise convolution layer in CFP will significantly increase the GPU memory cost, which consumes about 22GB (BasicVSR consumes about 13GB).
Due to the higher GPU memory cost, we can not train our model with GAN architecture on the RTX 3090s (24GB) server.
Thus, we do not compare our model with some GAN-based methods (\textit{e.g.,} \cite{yang2021real} and RealBasicVSR~\cite{chan2022investigating}).
However, the latest version of PyTorch ($\ge$1.12) is optimized for depth-wise convolution operations, which can alleviate the GPU memory cost.

\section{More Qualitative Results} \label{sec:visual}
\begin{figure}[!ht]
	\footnotesize
	\begin{center}
		\begin{tabular}{c c c c c c c c}
			\multicolumn{3}{c}{\multirow{5}*[34pt]{\hspace{-2mm} \includegraphics[width=0.49\linewidth, height=0.28\linewidth]{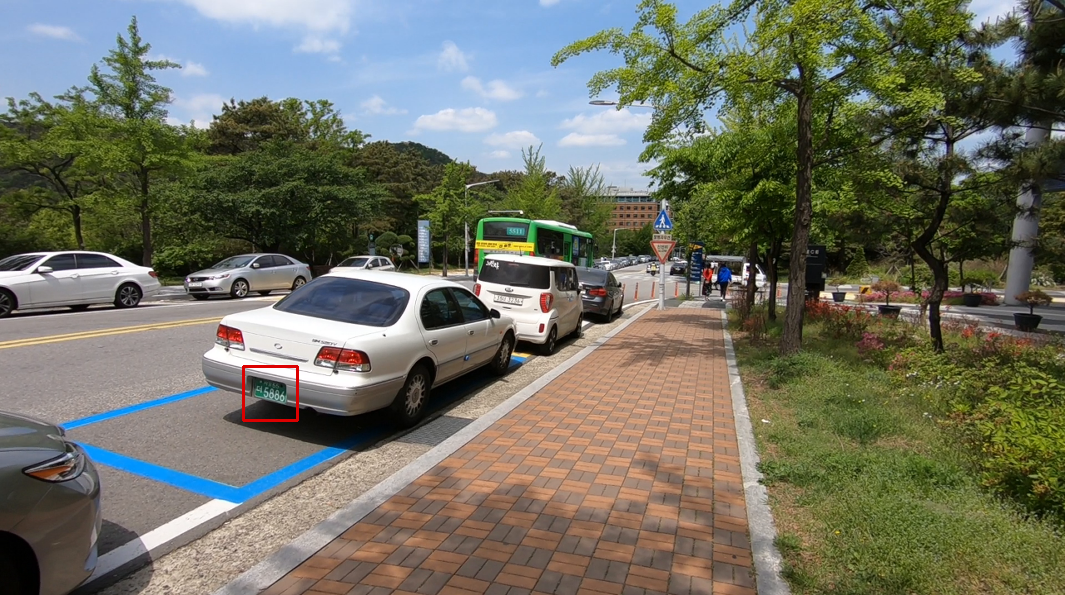}}}
            &  \includegraphics[width=0.12\linewidth]{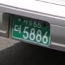}
            &  \includegraphics[width=0.12\linewidth]{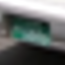}
            &   \includegraphics[width=0.12\linewidth]{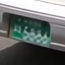}
            &  \includegraphics[width=0.12\linewidth]{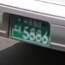}  \\
			\multicolumn{3}{c}{~}                                   & (a)                 &  (b)                   &  (c)  &  (d)  \\		
			\multicolumn{3}{c}{~}
            &  \includegraphics[width=0.12\linewidth]{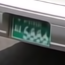}
            &  \includegraphics[width=0.12\linewidth]{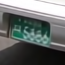}
            &  \includegraphics[width=0.12\linewidth]{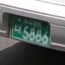}
            &  \includegraphics[width=0.12\linewidth]{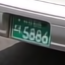}  \\
			\multicolumn{3}{c}{Frame 046, Clip 000} & (e)  &  (f)   &  (g)           & (h) \\			
		\end{tabular}
	\end{center}
	\vspace{-6mm}
	\caption{Visual comparisons on the REDS4~\cite{nah2019ntire} dataset. (a) Ground truth. (b)-(h) denote the results generated by Bicubic, BasicVSR~\cite{chan2021basicvsr}, BasicVSR++~\cite{chan2022basicvsrplusplus}, VRT~\cite{liang2022vrt}, TTVSR~\cite{liu2022ttvsr}, CFD-BasicVSR (Ours), and CFD-BasicVSR++ (Ours), respectively. Benefit from our proposed method, CFD-BaiscVSR has a similar visual quality result with BasicVSR++~\cite{chan2022basicvsrplusplus}. Moreover, our CFD-BasicVSR++ achieves a better reconstruction on the text of the license plate compared to other methods.}
	\label{fig:visual3}
	\vspace{-4mm}
\end{figure}

\begin{figure}[!ht]
	\footnotesize
	\begin{center}
		\begin{tabular}{c c c c c c c c}
			\multicolumn{3}{c}{\multirow{5}*[34pt]{\hspace{-2mm} \includegraphics[width=0.49\linewidth, height=0.28\linewidth]{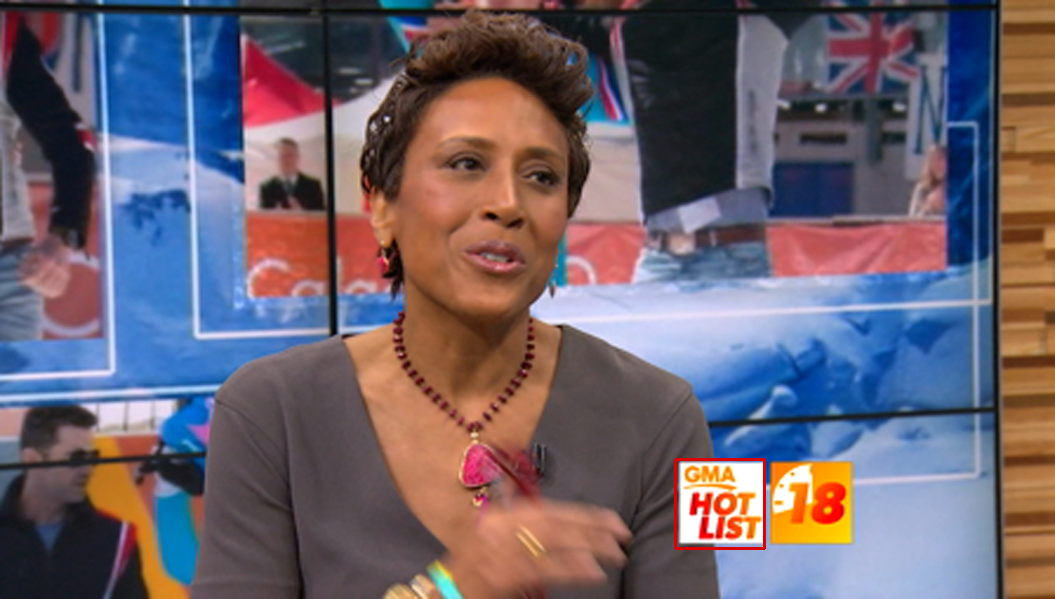}}}
            &  \includegraphics[width=0.12\linewidth]{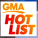}
            &  \includegraphics[width=0.12\linewidth]{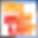}
            &   \includegraphics[width=0.12\linewidth]{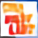}
            &  \includegraphics[width=0.12\linewidth]{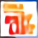}  \\
			\multicolumn{3}{c}{~}                                   & (a)                 &  (b)                   &  (c)  &  (d)  \\		
			\multicolumn{3}{c}{~}
            &  \includegraphics[width=0.12\linewidth]{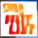}
            &  \includegraphics[width=0.12\linewidth]{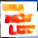}
            &  \includegraphics[width=0.12\linewidth]{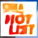}
            &  \includegraphics[width=0.12\linewidth]{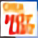}  \\
			\multicolumn{3}{c}{Sequence 829, Clip 003} & (e)  &  (f)   &  (g)           & (h) \\			
		\end{tabular}
	\end{center}
	\vspace{-6mm}
	\caption{Visual comparisons on the Vimeo-T~\cite{xue2019video} dataset. (a) Ground truth. (b)-(h) denote the results generated by Bicubic, BasicVSR~\cite{chan2021basicvsr}, BasicVSR++~\cite{chan2022basicvsrplusplus}, VRT~\cite{liang2022vrt}, TTVSR~\cite{liu2022ttvsr}, CFD-BasicVSR (Ours), and CFD-BasicVSR++ (Ours), respectively. All previous methods show different levels of artifacts due to the structures and edge details destroyed by feature alignment. However, our methods (g) and (h) can effectively reduce the influences of artifacts and restore clearer details of the image.}
	\label{fig:visual4}
	\vspace{-4mm}
\end{figure}

\begin{figure}[!ht]
	\footnotesize
	\begin{center}
		\begin{tabular}{c c c c c c c c}
			\multicolumn{3}{c}{\multirow{5}*[34pt]{\hspace{-2mm} \includegraphics[width=0.49\linewidth, height=0.28\linewidth]{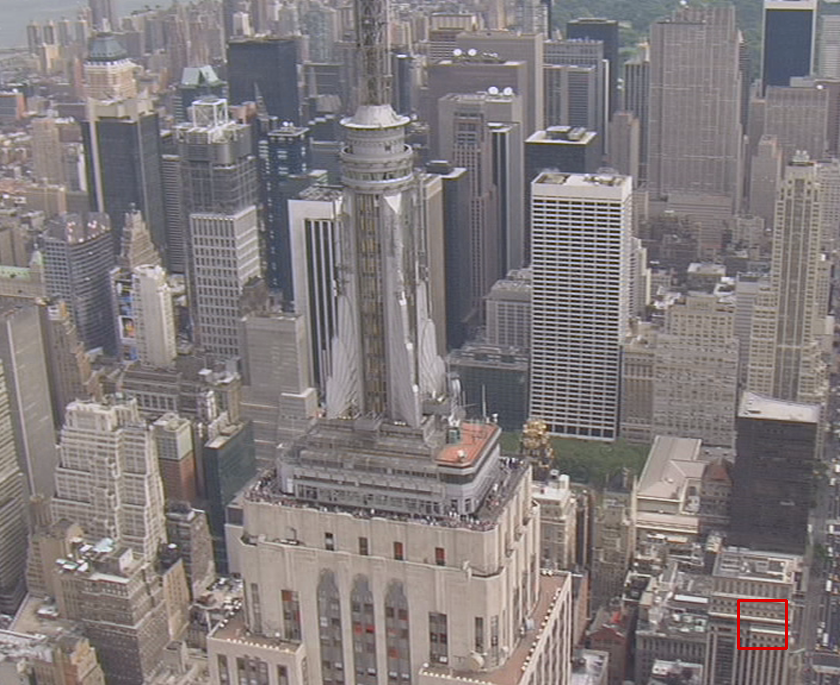}}}
            &  \includegraphics[width=0.12\linewidth]{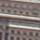}
            &  \includegraphics[width=0.12\linewidth]{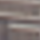}
            &   \includegraphics[width=0.12\linewidth]{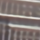}
            &  \includegraphics[width=0.12\linewidth]{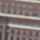}  \\
			\multicolumn{3}{c}{~}                                   & (a)                 &  (b)                   &  (c)  &  (d)  \\		
			\multicolumn{3}{c}{~}
            &  \includegraphics[width=0.12\linewidth]{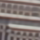}
            &  \includegraphics[width=0.12\linewidth]{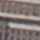}
            &  \includegraphics[width=0.12\linewidth]{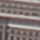}
            &  \includegraphics[width=0.12\linewidth]{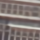}  \\
			\multicolumn{3}{c}{Frame 033, Clip city} & (e)  &  (f)   &  (g)           & (h) \\			
		\end{tabular}
	\end{center}
	\vspace{-6mm}
	\caption{Visual comparisons on the Vid4~\cite{liu2014bayesian} dataset. (a) Ground truth. (b)-(h) denote the results generated by Bicubic, BasicVSR~\cite{chan2021basicvsr}, BasicVSR++~\cite{chan2022basicvsrplusplus}, VRT~\cite{liang2022vrt}, TTVSR~\cite{liu2022ttvsr}, CFD-BasicVSR (Ours), and CFD-BasicVSR++ (Ours), respectively. BasicVSR~\cite{chan2021basicvsr}, BasicVSR++~\cite{chan2022basicvsrplusplus} and TTVSR~\cite{liu2022ttvsr} produce blurry outputs. Our methods (g) and (h) reconstruct more edge details of the windows compared to VRT~\cite{liang2022vrt}.}
	\label{fig:visual5}
	\vspace{-4mm}
\end{figure}

\begin{figure}[!ht]
	\footnotesize
	\begin{center}
		\begin{tabular}{c c c c c c c c}
			\multicolumn{3}{c}{\multirow{5}*[34pt]{\hspace{-2mm} \includegraphics[width=0.49\linewidth, height=0.28\linewidth]{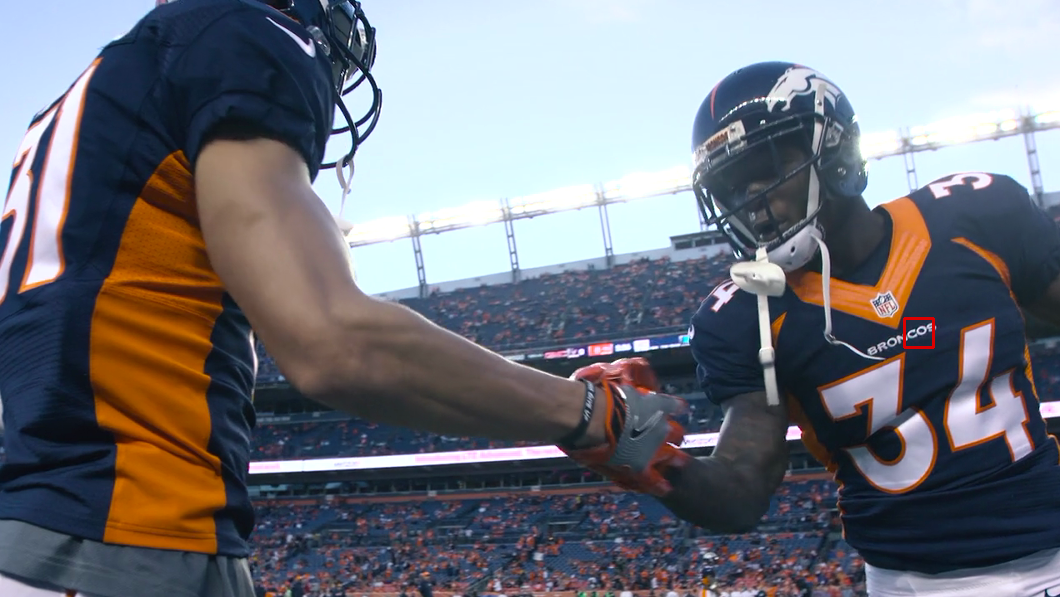}}}
            &  \includegraphics[width=0.12\linewidth]{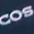}
            &  \includegraphics[width=0.12\linewidth]{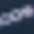}
            &   \includegraphics[width=0.12\linewidth]{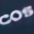}
            &  \includegraphics[width=0.12\linewidth]{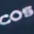}  \\
			\multicolumn{3}{c}{~}                                   & (a)                 &  (b)                   &  (c)  &  (d)  \\		
			\multicolumn{3}{c}{~}
            &  \includegraphics[width=0.12\linewidth]{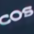}
            &  \includegraphics[width=0.12\linewidth]{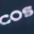}
            &  \includegraphics[width=0.12\linewidth]{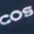}
            &  \includegraphics[width=0.12\linewidth]{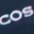}  \\
			\multicolumn{3}{c}{Frame 024, Clip clap} & (e)  &  (f)   &  (g)           & (h) \\			
		\end{tabular}
	\end{center}
	\vspace{-6mm}
	\caption{Visual comparisons on the UDM10~\cite{yi2019progressive} dataset. (a) Ground truth. (b)-(h) denote the results generated by Bicubic, BasicVSR~\cite{chan2021basicvsr}, BasicVSR++~\cite{chan2022basicvsrplusplus}, VRT~\cite{liang2022vrt}, TTVSR~\cite{liu2022ttvsr}, CFD-BasicVSR (Ours), and CFD-BasicVSR++ (Ours), respectively. The results in (c), (d), and (f) have some artifacts on the text `S'. Though the result (e) by VRT~\cite{liang2022vrt} has a better quality result, it produces some noises between the `O' and `S'. In contrast, our methods achieve a better reconstruction with finer details and structures.}
	\label{fig:visual6}
	\vspace{-4mm}
\end{figure}

In this section, we present additional visual comparisons with state-of-the-art methods~\cite{chan2021basicvsr,chan2022basicvsrplusplus,liang2022vrt,liu2022ttvsr} on the REDS4, Vimeo-T, Vid4, and UDM10 datasets.
Figure~\ref{fig:visual3}-\ref{fig:visual6} show that our models generate clearer images with finer details and structures than those by state-of-the-art methods.

\begin{figure*}[!htbp]
	
	\centering
	\tiny

	\newcommand{\wb}{0.16}
	\renewcommand{\tabcolsep}{0.8pt}
	\renewcommand{\arraystretch}{1}
	\resizebox{1.00\linewidth}{!} {
		\begin{tabular}{cc}
			
			\begin{tabular}{cc}
				\begin{adjustbox}{valign=t}
					\begin{tabular}{c}%
		         	\includegraphics[width=0.5\linewidth,height=0.383\linewidth]{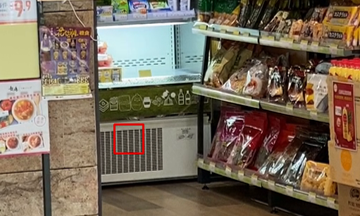}
						\\
						Frame 015, Clip 407
					\end{tabular}
				\end{adjustbox}
				\begin{adjustbox}{valign=t}
					\begin{tabular}{ccccc}
						\includegraphics[width=0.18 \textwidth]{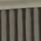} \hspace{0mm} &
						\includegraphics[width=0.18 \textwidth]{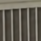} \hspace{0mm} &
						\includegraphics[width=0.18 \textwidth]{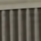}
						\\
						(a) BasicVSR~\cite{chan2021basicvsr} &
						(b) BasicVSR++~\cite{chan2022basicvsrplusplus} & (c) VRT~\cite{liang2022vrt}
						\\
						\vspace{-2.1mm}
						\\
						\includegraphics[width=0.18 \textwidth]{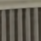} \hspace{0mm} &
						\includegraphics[width=0.18 \textwidth]{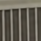}
						\hspace{0mm} &		
						\includegraphics[width=0.18 \textwidth]{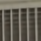}
						\\
						(d) TTVSR~\cite{liu2022ttvsr}  \hspace{0mm} &	(e) CFD-BasicVSR \hspace{0mm} &
						(f) CFD-BasicVSR++
						\\
					\end{tabular}
				\end{adjustbox}
			\end{tabular}
			
		\end{tabular}
	}	
	\vspace{-2mm}
	\caption{Visual quality comparison on the real-world VSR dataset~\cite{yang2021real}. Obviously, only our CFD-BasicVSR++ can restore the accurate edge details of the windows, demonstrating the generalization of our method in real scenes.} %
	\vspace{-4mm}
	\label{fig:visual7}
\end{figure*}

\begin{figure*}[!htbp]
	
	\centering
	\tiny

	\newcommand{\wb}{0.16}
	\renewcommand{\tabcolsep}{0.8pt}
	\renewcommand{\arraystretch}{1}
	\resizebox{1.00\linewidth}{!} {
		\begin{tabular}{cc}
			
			\begin{tabular}{cc}
				\begin{adjustbox}{valign=t}
					\begin{tabular}{c}%
		         	\includegraphics[width=0.5\linewidth,height=0.383\linewidth]{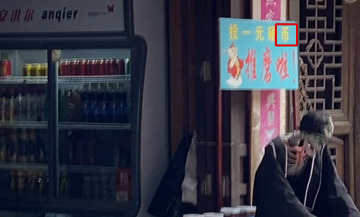}
						\\
						Frame 020, Clip 170
					\end{tabular}
				\end{adjustbox}
				\begin{adjustbox}{valign=t}
					\begin{tabular}{ccccc}
						\rotatebox{90}{\includegraphics[width=0.18 \textwidth]{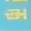}} \hspace{0mm} &
						\rotatebox{90}{\includegraphics[width=0.18 \textwidth]{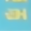}} \hspace{0mm} &
						\rotatebox{90}{\includegraphics[width=0.18 \textwidth]{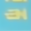}}
						\\
						(a) BasicVSR~\cite{chan2021basicvsr} &
						(b) BasicVSR++~\cite{chan2022basicvsrplusplus} & (c) VRT~\cite{liang2022vrt}
						\\
						\vspace{-2.1mm}
						\\
						\rotatebox{90}{\includegraphics[width=0.18 \textwidth]{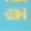}} \hspace{0mm} &
						\rotatebox{90}{\includegraphics[width=0.18 \textwidth]{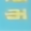}}
						\hspace{0mm} &		
						\rotatebox{90}{\includegraphics[width=0.18 \textwidth]{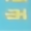}}
						\\
						(d) TTVSR~\cite{liu2022ttvsr}  \hspace{0mm} &	(e) CFD-BasicVSR \hspace{0mm} &
						(f) CFD-BasicVSR++
						\\
					\end{tabular}
				\end{adjustbox}
			\end{tabular}
			
		\end{tabular}
	}	
	\vspace{-2mm}
	\caption{Visual quality comparison on the real-world VSR dataset~\cite{yang2021real}. It shows that BasicVSR~\cite{chan2021basicvsr}, BasicVSR++~\cite{chan2022basicvsrplusplus} and TTVSR~\cite{liu2022ttvsr} generate different levels of artifacts on the text, while VRT~\cite{liang2022vrt} also has some blurring at the top of the image. However, due to the effectiveness of our DAC and CFP method, CFD-BasicVSR and CFD-BasicVSR++ achieve significant reconstruction improvement on the text of the real-world image.} %
	\label{fig:visual8}
\end{figure*}

Furthermore, to show the generalization of our models in real scenes, we use the real-world VSR dataset provided by~\cite{yang2021real} as the benchmark, and directly employ the pre-trained model of the REDS dataset to evaluate the SR performance.
Figure~\ref{fig:visual7} shows that CFD-BasicVSR++ reconstructs more edge details of windows than other methods.
Moreover, as shown in Figure~\ref{fig:visual8}, both our models restore clearer text structures.

\clearpage

\bibliographystyle{splncs04}
\bibliography{reference}